\documentclass[10pt,twocolumn,letterpaper]{article}
\PassOptionsToPackage{table,dvipsnames}{xcolor}
\PassOptionsToPackage{hyperfootnotes=false}{hyperref}
\usepackage[utf8]{inputenc}
\usepackage{amssymb}
\usepackage[round,authoryear]{natbib}
\usepackage{thunlp_cvpr}
\usepackage{ioa_layout}

\setthunlparticlemeta{RESEARCH ARTICLE}
\thunlplogoon
\title{Illusion of Alignment: Detecting Hidden Disagreement in Collaborative Dialogue}

\author{
 \textbf{Kaiming Liu\textsuperscript{1,2}},
 \textbf{Fuwen Luo\textsuperscript{2,3}},
 \textbf{Ziyue Wang\textsuperscript{2,3}},
 \textbf{Jinrui Ju\textsuperscript{3}}, \\
 \textbf{Yuxuan Liu\textsuperscript{3}}, 
 \textbf{Xuanyu Lei\textsuperscript{2,3}},
 \textbf{Yunghwei Lai\textsuperscript{2,3}},
 \textbf{Peng Li\textsuperscript{2,$\dagger$}},
 \textbf{Yang Liu\textsuperscript{1,2,3,$\dagger$}} \\
 \thunlpauthoraffildivider
}
\thunlpaffiliation{
\textsuperscript{1}College of AI, Tsinghua University, Beijing, China\\
 \textsuperscript{2}Institute for AI Industry Research (AIR), Tsinghua University, Beijing, China\\
 \textsuperscript{3}Dept. of Comp. Sci. \& Tech., Institute for AI, Tsinghua University, Beijing, China
}
\date{}

\begin{document}
\makefrontmatter{
Collaborative dialogue can end with apparent agreement while participants still differ on goals, assumptions, or execution plans, creating an \textbf{illusion of alignment (IoA)}.
A real-user study across 18 meetings confirms that IoA arises routinely in human collaboration.
Yet IoA poses a paradox: if participants were aware of such disagreements, they would already be explicit; if not, they cannot articulate them when asked, leaving IoA invisible to both participants and observers.
In this work, we make IoA detectable by generating diagnostic multiple-choice questions whose divergent answers across participants provide direct behavioral evidence of hidden disagreement.
We construct \textbf{IoA-Suite}, a dataset and evaluation protocol for detecting hidden disagreement, spanning five task types and six domains.
We find that even the best model attains only 49.5\% F1, with the bottleneck traced to private context that the dialogue does not surface.
We then train \textbf{IoA-Prober-8B} based on IoA-Suite, reaching 51.8\% F1 on IoA-Suite.
Across the aforementioned 18 real meetings, it surfaces 2.89 hidden disagreements per meeting that participants confirm they had not voiced, transferring to live human dialogue.
Further, in multi-agent collaboration, pairing IoA-Prober-8B with LLM agents improves downstream task performance on BigCodeBench-Hard and HiddenBench.
\ioaAbstractNote{Code, data and model weights are available at \url{https://github.com/THUNLP-MT/IoA}.}
}

\begingroup
\renewcommand{\thefootnote}{\fnsymbol{footnote}}
\footnotetext[2]{\raggedright\hangindent=1.8em\hangafter=1 Corresponding Authors: \href{mailto:lipeng@air.tsinghua.edu.cn}{lipeng@air.tsinghua.edu.cn}, \href{mailto:liuyang2011@tsinghua.edu.cn}{liuyang2011@tsinghua.edu.cn}.}
\endgroup

\section{Introduction}
\label{sec:introduction}
\begin{figure}[t]
  \centering
  \includegraphics[width=\linewidth]{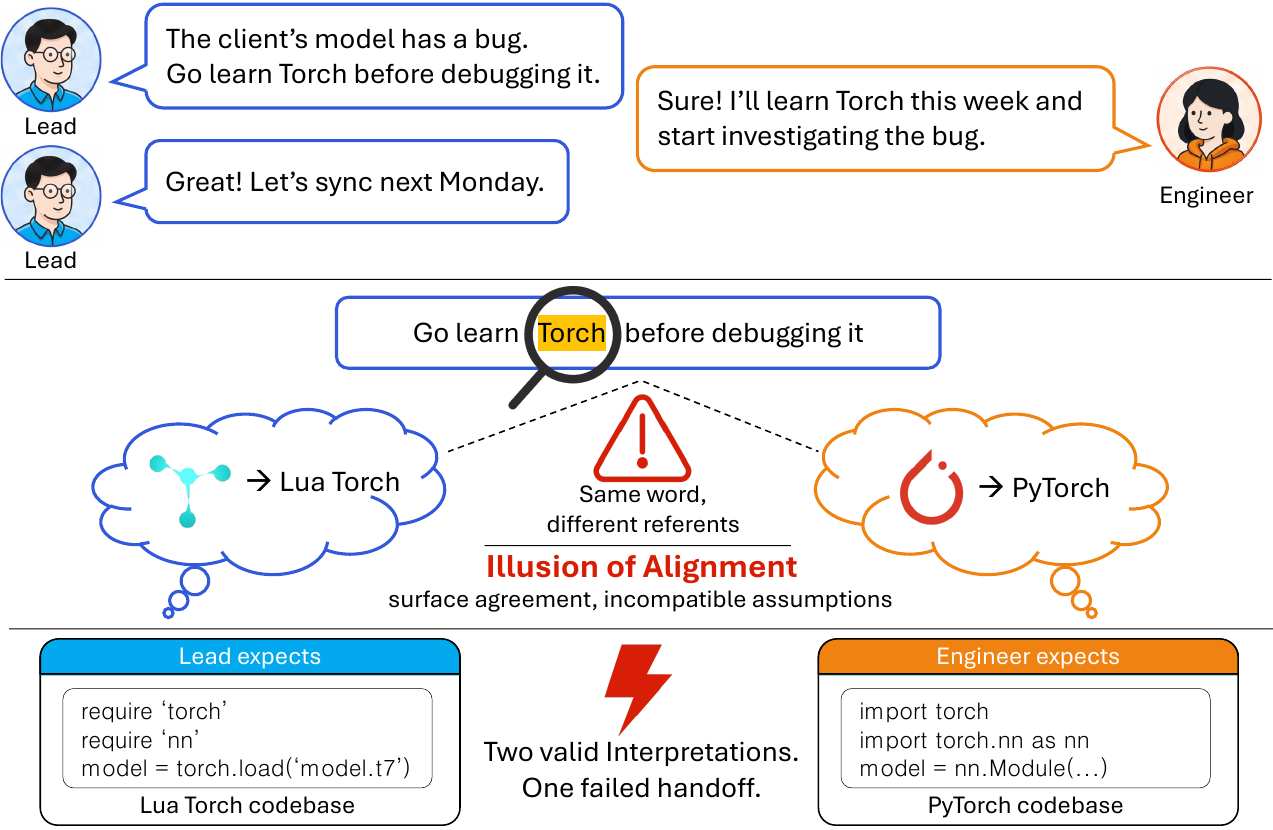}
  \caption{An example of the illusion of alignment. The lead and engineer appear to agree on the instruction ``\textit{Go learn Torch before debugging it}'', but the shared term hides incompatible assumptions: the lead refers to the legacy Lua-Torch codebase used by the client, whereas the engineer prepares for modern PyTorch. The task of this work is to surface such hidden misalignments before apparent consensus leads to a failed handoff.}
  \label{fig:teaser}
  \vspace{-8pt}
\end{figure}

Consider the scenario in Figure~\ref{fig:teaser}: a technical lead instructs an engineer to ``\textit{Go learn Torch before debugging it},'' and the engineer agrees.
A week later, the lead expects familiarity with a legacy Lua-Torch codebase, while the engineer has prepared around the modern PyTorch.
Neither party was careless, and to each participant the transcript contains no hedging, contradiction, or repair.
Each interpreted the shared term through a private context that the conversation never surfaced, yet this silent divergence ultimately derails the handoff.
We refer to this phenomenon as the \textbf{illusion of alignment (IoA)}: a collaborative dialogue ends in surface agreement while participants still differ on goals, assumptions, or execution plans.
In this work, \emph{alignment} refers to agreement among dialogue participants, rather than AI value alignment.

This phenomenon is far from a contrived edge case.
A real-user study (Section~\ref{sec:user-study}) across 18 meetings with 43 participants surfaces 2.89 hidden disagreements per meeting, which participants confirm they had not articulated during the original discussions.
While long recognized in organizational psychology as failures of shared mental models~\citep{cannon1993shared} and in management practice as a leading cause of execution breakdown~\citep{lencioni2002five}, IoA has received little attention in the computational dialogue research.

Beyond the scholarly neglect, detecting IoA poses a structural paradox.
Surface cues such as hedging or contradiction~\citep{de-kock-vlachos-2021-beg} are absent precisely because participants believe consensus has been reached, and existing work on dialogue disagreement studies only such explicit friction or observable common-ground failures~\citep{nath-etal-2025-frictional, sarkar-etal-2025-understanding}.
Detecting IoA therefore requires Theory-of-Mind reasoning~\citep{Premack_Woodruff_1978}: a detector must infer how different participants, given their public profiles, would map the same apparently aligned transcript to divergent interpretations, expectations, or downstream commitments.
This requirement differs from existing LLM Theory-of-Mind evaluations~\citep{wilf-etal-2024-think, chen-etal-2024-tombench}, which probe agent beliefs or knowledge at locations specified in advance by the evaluation; in IoA detection, no such locations are given, and identifying where participants diverge is itself the task.

Building on this perspective, we raise two research questions:
\emph{\textbf{RQ1}: How can hidden disagreement under illusion of alignment be elicited and evaluated in a verifiable manner?
\textbf{RQ2}: To what extent can task-specific training improve IoA detection from observable dialogue context?}

In this work, we recast IoA detection as a question generation task: given a dialogue, a detector generates a small set of diagnostic multiple-choice questions, and each participant answers them independently from their own perspective.
A question on which two participants select different options constitutes direct behavioral evidence of a hidden disagreement they did not surface during the conversation.
This reformulation relocates the IoA signal from the transcript, which by definition conceals it, to downstream participant behavior, turning an ill-posed introspection problem into a mechanically decidable filtering task.

To evaluate whether current models can probe \emph{accurately} (asking questions that genuinely reveal disagreements) and \emph{efficiently} (without burdening participants with noise), we construct \textbf{IoA-Suite}, the first dataset and evaluation protocol for IoA detection, with synthesized dialogues paired with ground-truth misalignment sets across five collaborative task types and six domains.
Evaluation across nine frontier models indicates that even the strongest model attains only 49.5\% F1, with further analysis tracing the bottleneck to the private context that the paradox of IoA conceals.
To bridge this gap, we train \textbf{IoA-Prober-8B} via an RL recipe, reaching 51.8\% F1, with the gains further validated in a real-user meeting study.
Furthermore, pairing IoA-Prober-8B with LLM agents improves downstream task performance in multi-agent collaboration.
Our contributions are summarized as follows:
\begin{itemize}[leftmargin=1.5em,itemsep=0pt,parsep=0.2em,topsep=0.1em,partopsep=0.0em]
    \item We bring the \textbf{illusion of alignment} into computational dialogue research, recasting hidden disagreement as a behavioral signal that can be elicited and measured directly through diagnostic multiple-choice questions.
    \item We construct \textbf{IoA-Suite} and show that IoA detection remains unsolved across current LLMs, with experiments tracing the difficulty to private context the dialogue does not surface.
    \item We train \textbf{IoA-Prober-8B}, a detector that surfaces hidden disagreements in real-user meetings and improves multi-agent collaboration.
\end{itemize}

\section{Related Work}
\label{sec:related_work}

\paragraph{Misalignment Detection in Dialogue.}
Prior work studies how participants lose track of shared understanding during a conversation.
Some approaches track common ground from the perspective of each participant, modeling what each speaker takes to be mutually accepted~\citep{markowska-etal-2023-finding, khebour-etal-2024-common, mohapatra-etal-2024-conversational, li-etal-2025-grounded}.
Others label conversational frictions, deliberation cues, or grounding rifts as signals of misalignment, sometimes drawing on gesture and prosody~\citep{10.1145/3610056, nath-etal-2024-thoughts, sarkar-etal-2025-understanding, shaikh-etal-2025-navigating, DBLP:journals/corr/abs-2501-17348, nath-etal-2025-frictional, vanderhoeven-etal-2025-trace}.
A separate thread treats disagreement as an explicit stance and detects it via classification~\citep{li-etal-2023-new, wagner2025the, 10.1016/j.eswa.2025.127790, DBLP:journals/corr/abs-2603-23531}.
What unites these efforts is that misalignment leaves a trace, such as a token, gesture, or hesitation that an annotator can point to.
IoA leaves no such trace, since the words on the surface look like agreement and there is nothing to flag.

\begin{figure*}[t]
    \centering
    \includegraphics[width=\textwidth]{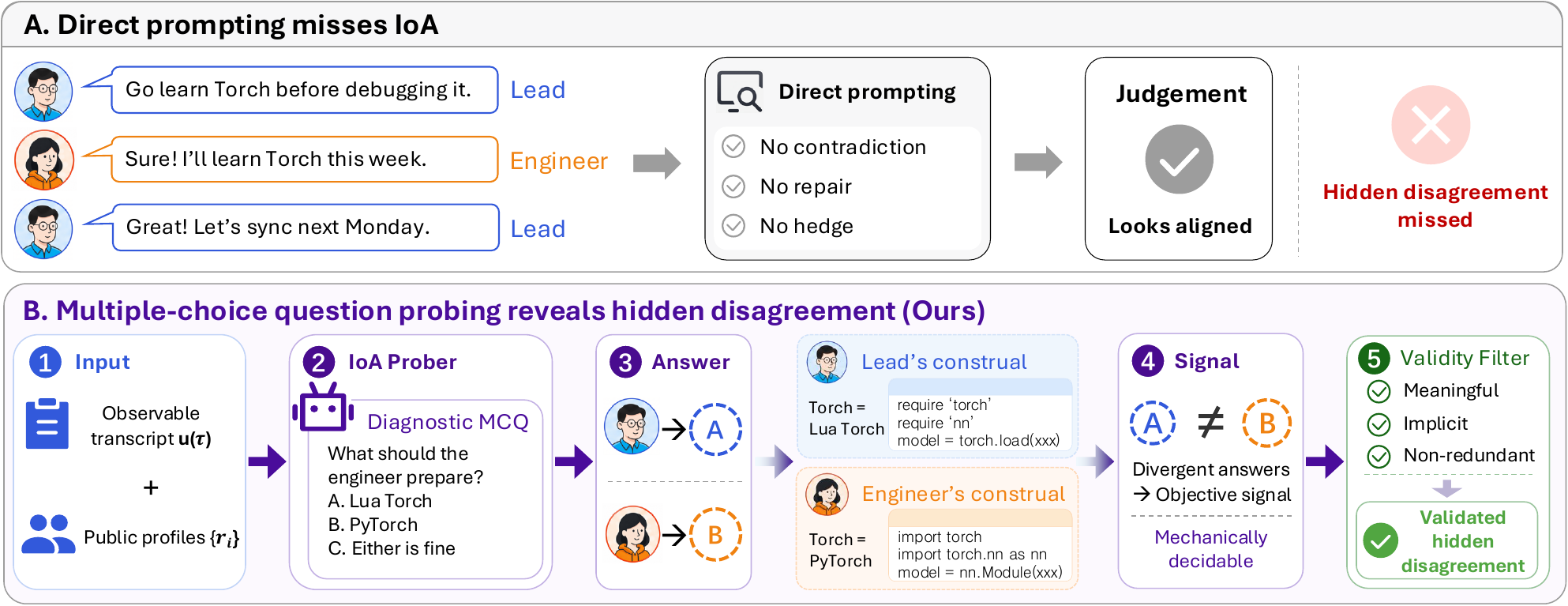}
    \caption{Overview of IoA detection through multiple-choice question (MCQ) probing.
    \textbf{(1)~Panel A} illustrates that direct prompting judges the dialogue as aligned when the transcript contains no contradiction, repair, or hedge, thereby missing latent disagreement.
    \textbf{(2)~Panel B} shows that our proposed IoA Prober generates MCQ from the observable transcript and public profiles.
    Independent answers expose divergent construals over the same dialogue.
    The validity filter is performed by participants in real-world deployment and automated in IoA-Suite for evaluation.
    }
    \label{fig:method}
\end{figure*}

\paragraph{Theory of Mind (ToM) in Dialogue.}
Work on ToM in LLMs falls into three groups.
Early benchmarks adapt the Sally-Anne task to text, asking models to follow character beliefs through short stories~\citep{le-etal-2019-revisiting, wu-etal-2023-hi, 10.5555/3666122.3666717, xu-etal-2024-opentom}.
Later work moves to dialogue, testing belief and knowledge inference under information asymmetry, multi-party exchange, or persuasion~\citep{bara-etal-2021-mindcraft, kim2023fantom, chen-etal-2024-tombench}.
A recent line places ToM inside user-agent interaction, where the agent must square latent user beliefs with task or environment states~\citep{qiu-etal-2024-minddial, jafari-etal-2025-beyond, ruan-etal-2026-beyond}.
Two assumptions run through all of them: the evaluation specifies where the model should attend, and answers are checked against an external ground truth.
IoA detection violates both: the divergence holds between participants rather than against an external truth, and identifying where it occurs is itself the task.

Taken together, IoA differs in three respects compared with prior work: the dialogue ends in apparent agreement with no explicit disagreement trace, the detector must identify and generate the probe itself, and disagreement is established behaviorally through divergent participant answers.

\section{Problem Formalization}
\label{sec:formulation}

To make illusion of alignment amenable to detection, we formalize it as a latent mismatch between how participants construe the same dialogue at its close.
The mismatch is latent because the transcript registers no friction, yet it determines whether participants would commit to incompatible follow-ups on issues at stake in the dialogue.

\paragraph{Participants.}
Consider a collaborative dialogue among $n \geq 2$ participants $P = \{p_1, \dots, p_n\}$.
Each $p_i$ is characterized by three layers of context.
A \emph{public profile} $r_i$ captures attributes observable to others, including role and declared expertise.
A \emph{private agenda} $g_i$ captures the goals and priorities $p_i$ pursues without articulating them in the meeting.
A set of \emph{tacit assumptions} $a_i$ captures the premises $p_i$ treats as already settled and therefore does not consider worth stating.
The triple $(r_i, g_i, a_i)$ shapes the utterances $p_i$ contributes and the commitments $p_i$ would later make.

\paragraph{Dialogues.}
Conditioned on the contexts of all participants, the conversation unfolds as a sequence of turns $\tau = (t_1, \dots, t_T)$, where at each turn $t_k$ a speaker $s_k \in P$ forms an inner thought $h_k$ and produces a publicly observable utterance $u_k$.
We write $u(\tau) = (u_1, \dots, u_T)$ for the observable transcript.

\paragraph{Illusion of Alignment.}
Each participant $p_i$ holds a \emph{post-dialogue construal} of what was agreed and what should follow, represented as
\begin{equation*}
c_i(\iota) = \mathcal{C}(\iota;\, r_i, g_i, a_i, u(\tau)),
\end{equation*}
where $\iota \in \mathcal{I}$ indexes the commitments at issue in the dialogue and $c_i(\iota) \in \mathcal{Y}$ is the commitment $p_i$ would make on $\iota$, jointly shaped by the three context layers of $p_i$ and the shared transcript.
The dialogue exhibits an \textbf{illusion of alignment} when $\tau$ ends in apparent agreement but two or more construals diverge on some issue.
The set of such divergence points,
\begin{equation*}
M(\tau) = \{\iota \in \mathcal{I} \mid \exists\, i \neq j,\ c_i(\iota) \neq c_j(\iota)\},
\end{equation*}
constitutes the latent disagreements on $\tau$.
Surfacing the elements of $M(\tau)$ in a verifiable manner is the detection task formulated in Section~\ref{sec:benchmark}.

\section{Detection and Evaluation: IoA-Suite}
\label{sec:benchmark}

We now address \emph{\textbf{RQ1}} by converting the latent disagreement defined in Section~\ref{sec:formulation} into a verifiable detection task.
We introduce \textbf{IoA-Suite}, a framework for IoA detection comprising three components: a detection method (Section~\ref{sec:detection}), an evaluation protocol (Section~\ref{sec:eval_protocol}), and a quality-controlled synthesized dataset (Section~\ref{sec:benchmark_construction}).

\subsection{Detection Method}
\label{sec:detection}

The central design constraint is to expose elements of $M(\tau)$ without any model deciding whether two participants really disagree.
A transcript-level surface detector cannot meet this constraint: when the dialogue ends in apparent agreement, the absence of contradiction, repair, or hedging is by construction the very condition under which IoA arises, so a detector built on these surface signals would conclude that the dialogue is aligned (Figure~\ref{fig:method}, Panel~A).
We instead anchor the disagreement label in observable participant behavior, restricting model role to proposing candidate questions that the participants themselves separate by answering.

\paragraph{Multiple-choice question (MCQ) probing.}
We propose MCQ probing, as illustrated in Figure~\ref{fig:method}, Panel~B.
An IoA Prober $\mathcal{P}$ generates a set of candidate diagnostic questions from public context,
\begin{equation*}
\mathcal{P}\bigl(u(\tau),\ \{r_i\}_{i=1}^n\bigr) = Q = \{q_1, \dots, q_K\},
\end{equation*}
mirroring an external observer who has only the transcript and the declared identities of the speakers.
Each participant then answers every $q \in Q$ from the perspective of $p_i$, yielding $\{c_i(q)\}_{i=1}^n$, and $q$ is divergent when $c_i(q) \neq c_j(q)$ for some $i \neq j$.
The multiple-choice form is essential here: it fixes the answer space in advance, so divergence reduces to a mechanical check of whether two participants selected different options, without requiring any external judgment to compare phrasings for substantive disagreement.
Disagreement existence is therefore never adjudicated by a model verdict, which makes IoA detection objectively verifiable.

\subsection{Evaluation Protocol}
\label{sec:eval_protocol}

The MCQ probing method above yields a divergence label per question, but not every divergent question reflects a substantive latent disagreement.
We therefore introduce a \emph{validity filter} $\mathcal{V}$ to automatically remove divergent questions that fail to constitute genuine hidden disagreements.
In real-world meetings, this role is played by the participants themselves (Section~\ref{sec:user-study}).
Algorithm~\ref{alg:eval_protocol} instantiates the full evaluation pipeline.

\paragraph{Participant simulation.}
Each construal $c_i$ is realized by a role-playing simulator $S$~\citep{DBLP:journals/nature/ShanahanMR23} given the full context of $p_i$ ($r_i, g_i, a_i$ and inner thoughts $h_i(\tau)$) and only the public profiles $\{r_j\}_{j \neq i}$ plus the shared transcript $u(\tau)$ for others, matching what $p_i$ actually holds during the conversation.
We invoke $S$ once per participant, since a single roll-out already matches the majority vote over repeated samples (Appendix~\ref{app:eval-simulator}).
We also test sensitivity to question and option ordering: after randomly permuting both, divergence outcomes remain consistent for over 92\% of questions and aggregate F1 changes only marginally (Table~\ref{tab:order_sensitivity}).

\paragraph{Validity filter.}
For each divergent question $q \in Q_{\text{div}}$, $\mathcal{V}$ checks three criteria: \emph{meaningfulness} (the answer divergence reflects a real cognitive gap rather than wording noise), \emph{implicitness} ($q$ does not target a point already explicitly debated in $\tau$), and \emph{non-redundancy} ($q$ does not restate a divergence already covered by an earlier question in $Q$).
If $q$ passes all three, $\mathcal{V}$ attempts to attribute it to some $m \in M(\tau)$.
Crucially, $\mathcal{V}$ never decides whether participants disagree, which is settled mechanically by simulator outputs.

\begin{algorithm}[t]
\caption{Evaluation Protocol}
\label{alg:eval_protocol}
\small
\begin{algorithmic}[1]
\Require $(\tau, M(\tau), \{r_i, g_i, a_i\}_{i=1}^n)$; IoA Prober $\mathcal{P}$; Simulator $S$; Validity filter $\mathcal{V}$
\Ensure $P$, $R$, $\mathrm{F1}$
\State $Q \gets \mathcal{P}(u(\tau),\, \{r_i\}_{i=1}^n)$
\For{$i = 1, \dots, n$}
    \State $\mathbf{o}_i \gets S(Q;\, r_i, g_i, a_i, h_i(\tau);\, \{r_j\}_{j \neq i};\, u(\tau))$
\EndFor
\State $Q_{\text{div}} \gets \{q_k \in Q : \neg\, \mathrm{all\text{-}equal}(\mathbf{o}_1[k], \dots, \mathbf{o}_n[k])\}$
\State $\{(\text{acc}_k, \text{hit}_k)\}_{q_k \in Q_{\text{div}}} \gets$
\Statex \hspace{1.5em} $\mathcal{V}(Q_{\text{div}},\, \{\mathbf{o}_i\},\, \tau,\, M(\tau),\, \{r_i, g_i, a_i\})$
\State $A, B, C \gets \emptyset$
\For{$q_k \in Q$}
    \If{$q_k \notin Q_{\text{div}} \lor \neg\, \text{acc}_k$}\ $C \gets C \cup \{q_k\}$
    \ElsIf{$\text{hit}_k \in M(\tau)$}\ $A \gets A \cup \{(q_k, \text{hit}_k)\}$
    \Else\ $B \gets B \cup \{q_k\}$
    \EndIf
\EndFor
\State $P \gets |A| / (|A| + |C|)$
\State $R \gets |\{m \in M(\tau) : \exists\, (q, m) \in A\}| / |M(\tau)|$
\State \Return $(P,\, R,\, 2PR/(P+R))$
\end{algorithmic}
\end{algorithm}

\paragraph{Metrics.}
Each $q$ falls into category $A$ (passes $\mathcal{V}$ and attributed), $B$ (passes $\mathcal{V}$ but unattributed), or $C$ (non-divergent or rejected).
$B$ arises because dialogue generation may introduce emergent misalignments beyond the planted $M(\tau)$, which are real that $\mathcal{V}$ accepts but they have no ground-truth $m$ to attribute against.
We treat IoA-Suite F1 as a controlled synthetic proxy for IoA detection capability, rather than a complete measure of hidden disagreement in real collaboration.
We exclude $B$ from precision and recall to keep the score stable, and define F1 as the harmonic mean of $P$ and $R$.
$|B|$ is reported separately in Appendix~\ref{app:prober-full-results} as a complementary axis tracking divergences beyond the planted set.
The simulator uses Gemini-3-Flash~\citep{gemini3} and the judge uses GPT-5.4~\citep{gpt5.4}.
Full prompts and settings are in Appendix~\ref{app:eval}.

\subsection{Dataset Construction}
\label{sec:benchmark_construction}

Since the latent disagreement set $M(\tau)$ cannot be recovered from naturally occurring transcripts, we synthesize the dialogues in IoA-Suite by planting $M(\tau)$ before $\tau$ is generated, so that ground truth persists by construction.

\paragraph{Pipeline.}
Algorithm~\ref{alg:bench_construction} summarizes the two-stage procedure with stage-wise filtering.
Stage~1 instantiates two participants together with a planted $M(\tau)$, at least one element of which is \emph{definitional}, anchored to a shared term that participants use while privately referring to different things; this is the form of IoA hardest to surface with transcript-level signals and matches the canonical example of Figure~\ref{fig:teaser}.
Stage~2 generates $\tau$ under a hard constraint forbidding any utterance from explicitly debating $M(\tau)$, keeping the planted disagreements latent.
After each stage, a judge scores its output on stage-specific dimensions and routes failures to one round of regeneration before discarding.
Both generator and judge are GPT-5.4~\citep{gpt5.4}; full rubrics and prompts are in Appendix~\ref{app:ioa-suite}.

\begin{algorithm}[t]
\caption{IoA-Suite Dataset Construction}
\label{alg:bench_construction}
\small
\begin{algorithmic}[1]
\Require Seed $(d, k, x)$; Generator $G$; Judge $J$; Thresholds $\theta_1, \theta_2$
\Ensure Dialogue $\tau$ with Ground-truth $M(\tau)$, or \textsc{Discard}
\Statex \textit{// Stage 1: participants and planted misalignments}
\State $(\{r_i, g_i, a_i\}_{i=1}^n,\, M(\tau)) \gets G_{\text{participants}}(d, k, x)$
\State $s_1 \gets J_{\text{participants}}(\{r_i, g_i, a_i\}, M(\tau))$
\If{$s_1 < \theta_1$}
    \State regenerate once with feedback; update $s_1$
    \State \textbf{if} $s_1 < \theta_1$ \textbf{then return} \textsc{Discard}
\EndIf
\Statex \textit{// Stage 2: dialogue generation under latent constraint}
\State $\tau \gets G_{\text{dialogue}}(\{r_i, g_i, a_i\}, M(\tau))$
\State $s_2 \gets J_{\text{dialogue}}(\tau, \{r_i, g_i, a_i\}, M(\tau))$
\If{$s_2 < \theta_2$}
    \State regenerate once with feedback; update $s_2$
    \State \textbf{if} $s_2 < \theta_2$ \textbf{then return} \textsc{Discard}
\EndIf
\State \Return $(\tau, M(\tau), \{r_i, g_i, a_i\})$
\end{algorithmic}
\end{algorithm}

\paragraph{Seeds and coverage.}
Each dialogue is grounded in a real-world collaborative scenario specified by a seed tuple $(d, k, x)$, where $(d, k)$ specifies the domain and task type, and $x$ provides the substantive content for participants to discuss.
We instantiate $x$ as a scholarly publication, which anchors each dialogue in a concrete work-relevant scenario with enough substantive content to support genuine divergent interpretations rather than generic role-play.
We draw from five task types in the Group Task Circumplex~\citep{mcgrath1984groups}, namely \emph{planning}, \emph{design review}, \emph{troubleshooting}, \emph{coordination}, and \emph{brainstorming}, paired with six domains: ML research, software engineering, cross-disciplinary work, medicine, business, and law.
Train, validation, and test splits draw from disjoint seed pools, eliminating leakage at the level of source material.

\paragraph{Statistics and quality.}
The procedure yields 1{,}200 training, 120 validation, and 300 test dialogues.

Since GPT-5.4 supplies both generator and judge, self-enhancement bias~\citep{10.5555/3666122.3668142, 10.5555/3737916.3740113} is a concern, which we address with two complementary external checks on the test set: re-scoring by Gemini-3.1-Pro Preview~\citep{gemini3.1}, and parallel evaluation by two trained PhD annotators.
As illustrated in Appendix~\ref{app:human_eval}, both the cross-family automated re-scoring and independent human annotation corroborate the quality of IoA-Suite.

\section{IoA Detection Bottleneck Diagnostic}
\label{sec:diagnosis}

With IoA-Suite in place, we ask how current models perform on IoA detection, and trace the bottleneck to the information they have access to.

\subsection{IoA Detection Is Unsolved at the Frontier}
\label{sec:bench_results}

\begin{table}[t]
\centering
\small
\renewcommand{\arraystretch}{1.15}
\resizebox{\columnwidth}{!}{%
\begin{tabular}{@{}lccc!{\color{gray!50}\vrule}ccc@{}}
\toprule
& \multicolumn{3}{c!{\color{gray!50}\vrule}}{\textit{Direct prompting}} & \multicolumn{3}{c}{\textit{MCQ probing (ours)}} \\
\cmidrule(lr){2-4} \cmidrule(lr){5-7}
\textbf{Model} & \textbf{P} & \textbf{R} & \textbf{F1} & \textbf{P} & \textbf{R} & \textbf{F1} \\
\midrule
\rowcolor{gray!4}\multicolumn{7}{@{}l}{\textit{Closed-source}} \\
\midrule
GPT-5.4                & 29.6 & 36.1 & 32.2 & 44.7 & \textbf{57.7} & \textbf{49.5}\,\textcolor{green!50!black}{\scriptsize(+17.3)} \\
Gemini-3.1-Pro$^{\dagger}$         & 35.8 & 32.4 & 33.8 & \textbf{46.5} & 40.4 & 43.0\,\textcolor{green!50!black}{\scriptsize(+\phantom{0}9.2)} \\
Gemini-3-Flash         & 24.3 & 27.1 & 25.3 & 43.9 & 40.4 & 41.9\,\textcolor{green!50!black}{\scriptsize(+16.6)} \\
GPT-5 mini             & 20.3 & 36.0 & 25.8 & 34.9 & 47.8 & 40.0\,\textcolor{green!50!black}{\scriptsize(+14.2)} \\
\midrule
\rowcolor{gray!4}\multicolumn{7}{@{}l}{\textit{Open-source}} \\
\midrule
DeepSeek-V3.2          & 22.7 & 28.2 & 25.0 & 42.8 & \underline{43.6} & \underline{42.9}\,\textcolor{green!50!black}{\scriptsize(+17.9)} \\
GLM-5.1                & 24.7 & 29.5 & 26.6 & 41.3 & 38.6 & 39.3\,\textcolor{green!50!black}{\scriptsize(+12.7)} \\
Qwen3.5-397B$^{\dagger}$      & 21.7 & 23.5 & 22.4 & \underline{46.3} & 34.6 & 38.1\,\textcolor{green!50!black}{\scriptsize(+15.7)} \\
Kimi-K2.5              & 20.9 & 31.1 & 24.9 & 35.5 & 39.8 & 37.1\,\textcolor{green!50!black}{\scriptsize(+12.2)} \\
Qwen3-8B               & 23.3  & 21.7  & 22.3  & 37.5 & 23.3 & 28.0\,\textcolor{green!50!black}{\scriptsize(+\phantom{0}5.7)} \\
\bottomrule
\end{tabular}
}
\caption{Performance (\%) on IoA-Suite test split under direct prompting and MCQ probing (our protocol from Section~\ref{sec:detection}).
The green annotation next to each MCQ probing F1 reports the gain over the direct prompting baseline on the same model.
\textbf{Bold}: best overall.
\underline{Underline}: best within open-source.
$^{\dagger}$Abbreviated names: Gemini-3.1-Pro for Gemini-3.1-Pro Preview, Qwen3.5-397B for Qwen3.5-397B-A17B; the same abbreviations apply throughout the paper.}
\label{tab:main_results}
\vspace{-8pt}
\end{table}

We evaluate four closed-source and five open-source models on IoA-Suite under two protocols: \emph{direct prompting}, where the model is asked to list hidden disagreements in the dialogue, and \emph{MCQ probing}, our protocol from Section~\ref{sec:detection}.
Results appear in Table~\ref{tab:main_results}; full prompts, results, hyperparameters are in Appendix~\ref{app:eval}.

\paragraph{MCQ probing is effective for IoA detection.}
Every model gains substantially from the MCQ mechanism: F1 improves by 5.7\% to 17.9\% across the nine models tested under both protocols, with a mean gain of 13.5\%.
Crucially, recall also rises in every case (mean +11.2\%), so the gain is not driven by inflated question counts.
The pattern confirms the design rationale of Section~\ref{sec:detection}: anchoring detection in participant behavior, rather than asking a model to declare disagreements directly, produces a more reliable signal for IoA.

\paragraph{A low ceiling across families.}
No model exceeds 50.0\% F1 in IoA-Suite, with the strongest reaching only 49.5\%.
The gap between closed- and open-source families is narrow: DeepSeek-V3.2 trails GPT-5.4 by just 6.6\% F1, far less than typical reasoning-benchmark gaps.
The low ceiling and narrow spread suggest that IoA detection is a capability current models broadly lack, and one unlikely to emerge from general reasoning scaling.

\paragraph{Question-asking strategies diverge across families.}
Beyond aggregate F1, the average number of questions generated per dialogue ($K$ in Table~\ref{tab:full-results}) under MCQ probing reveals two opposing strategies.
GPT-series models probe broadly ($K = 5.8$ for both GPT-5.4 and GPT-5 mini), trading precision for recall; Gemini-series models probe selectively ($K = 3.5\sim3.8$) at higher precision but lower recall.
This trade-off is inherent to prompting alone: neither family achieves high recall without inflating $K$, motivating an objective that directly optimizes both axes (Section~\ref{sec:training}).

\paragraph{IoA is not reducible to lexical ambiguity.}
To examine whether IoA detection can be reduced to lexical ambiguity, we conduct two analyses, with full results reported in Appendix~\ref{app:lexical}: a taxonomy analysis over IoA-Suite and a comparison with lexical ambiguity baselines.
Definitional disagreements account for only 24.2\% of IoA-Suite, while criterial, scopal, presuppositional, and procedural disagreements constitute the majority.
Moreover, both a WordNet lexical baseline and stronger LLM ranked ambiguity baselines substantially underperform direct MCQ probing, indicating that effective probing requires inference over distinct construals of the interlocutors rather than simple identification of ambiguous expressions.

\subsection{Private Context Matters}
\label{sec:input_ablation}

\begin{table}[t]
\centering
\small
\renewcommand{\arraystretch}{1.15}
\resizebox{.95\columnwidth}{!}{%
\begin{tabular}{@{}llccc@{}}
\toprule
\textbf{Model} & \textbf{Setting} & \textbf{P} & \textbf{R} & \textbf{F1}\phantom{\,\scriptsize(+23.8)} \\
\midrule
\multirow{4}{*}{GPT-5.4}
 & \cellcolor{gray!8}default        & \cellcolor{gray!8}44.7 & \cellcolor{gray!8}57.7 & \cellcolor{gray!8}49.5\phantom{\,\scriptsize(+23.8)} \\
 & \quad + $h_k$                                  & 72.0 & 77.2 & 74.1\,\textcolor{green!50!black}{\scriptsize(+24.6)} \\
 & \quad + $g_i, a_i$                             & 80.3 & 84.3 & 82.0\,\textcolor{green!50!black}{\scriptsize(+32.5)} \\
 & \quad + $g_i, a_i, h_k$                        & 85.3 & 83.9 & 84.4\,\textcolor{green!50!black}{\scriptsize(+34.9)} \\
\midrule
\multirow{4}{*}{Gemini-3.1-Pro$^{\dagger}$}
 & \cellcolor{gray!8} default        & \cellcolor{gray!8}46.5 & \cellcolor{gray!8}40.4 & \cellcolor{gray!8}43.0\phantom{\,\scriptsize(+23.8)} \\
 & \quad + $h_k$                                  & 80.0 & 72.2 & 75.4\,\textcolor{green!50!black}{\scriptsize(+32.4)} \\
 & \quad + $g_i, a_i$                             & 83.1 & 81.4 & 82.0\,\textcolor{green!50!black}{\scriptsize(+39.0)} \\
 & \quad + $g_i, a_i, h_k$                        & 80.0 & 85.0 & 82.1\,\textcolor{green!50!black}{\scriptsize(+39.1)} \\
\midrule
\multirow{4}{*}{Qwen3-8B}
 & \cellcolor{gray!8} default        & \cellcolor{gray!8}37.5 & \cellcolor{gray!8}23.3 & \cellcolor{gray!8}28.0\phantom{\,\scriptsize(+23.8)} \\
 & \quad + $h_k$                                  & 62.0 & 44.4 & 50.9\,\textcolor{green!50!black}{\scriptsize(+22.9)} \\
 & \quad + $g_i, a_i$                             & 81.8 & 72.3 & 75.8\,\textcolor{green!50!black}{\scriptsize(+47.8)} \\
 & \quad + $g_i, a_i, h_k$                        & 80.7 & 73.4 & 76.4\,\textcolor{green!50!black}{\scriptsize(+48.4)} \\
\bottomrule
\end{tabular}
}
\caption{Effect of supplying oracle private context to the prober.
The \colorbox{gray!8}{default} row exposes only $r_i$ and $u(\tau)$, mirroring the deployment setting evaluated in Section~\ref{sec:bench_results}.
Subsequent rows additionally expose turn-level inner thoughts $h_k$, private agenda and tacit assumptions $(g_i, a_i)$, or both.
Green annotations report the gain over the default for the same model 
(all values in \%).}
\label{tab:input_ablation}
\vspace{-8pt}
\end{table}

To localize the bottleneck, we lift the default $r_i$-only restriction along two axes: turn-level inner thoughts $h_k$ and static private context $(g_i, a_i)$, evaluating GPT-5.4, Gemini-3.1-Pro Preview, and Qwen3-8B~\citep{qwen3} under the resulting four settings (Table~\ref{tab:input_ablation}; full prompts in Appendix~\ref{app:prober-ablation}).

Either axis of private context alone is enough to transform performance, lifting F1 by over 20.0\% absolute across all three models, while combining both yields only marginal further gains.
This redundancy indicates that the two axes signal the same underlying quantity: the private perspective each participant holds on the conversation.
Once the private context is available, the three models converge to a narrow band (75.8\%$\sim$84.4\%), with Qwen3-8B exceeding every model in the default setting.

Two implications follow.
First, IoA detection is fundamentally a Theory-of-Mind task: it requires recovering the private perspective each participant holds rather than reasoning more carefully over what is already said, which reframes the training objective in Section~\ref{sec:training} as learning to infer this perspective from the public transcript alone.
Second, IoA-Suite is well-calibrated: it stays hard under the realistic public-only input yet becomes tractable once oracle private context is supplied, confirming that the gap reflects information access rather than an artifact of the construction.

\section{Improving IoA Detection: IoA-Prober}
\label{sec:training}

The results in Section~\ref{sec:diagnosis} show that IoA detection remains difficult when only public context is available.
We address \emph{\textbf{RQ2}} by training a prober directly on the diagnostic question generation objective.

\subsection{Training Recipe and Main Results}
\label{sec:prober_main}

\paragraph{Training recipe.}
IoA-Prober-8B is trained from Qwen3-8B in two stages.
The first is a supervised warm-start on 300 dialogues sampled from the training split: for each dialogue we collect candidate questions from GPT-5.4 and resolve through the pipeline of Section~\ref{sec:eval_protocol}, retaining Categories A and B (divergent questions accepted by the validity filter, with and without ground-truth attribution respectively) and dropping Category C (non-divergent or filter-rejected questions) which would push the model toward uninformative outputs.
The second stage applies GRPO~\citep{deepseek-math} with F1 as the reward, sampling a group of candidate question sets per dialogue and scoring each end-to-end through the same pipeline:
\begin{equation*}
\resizebox{\columnwidth}{!}{$
\displaystyle
\mathcal{J}_{\text{GRPO}}(\pi_\theta)
=
\mathbb{E}_{\tau \sim \mathcal{D},\,\{Q^j\}_{j=1}^G \sim \pi_\theta}
\!\left[
\frac{1}{G}\sum_{j=1}^{G}
A_j \log \pi_\theta(Q^j \mid \tau)
\right]
$}
\end{equation*}
where $A_j$ is the group-normalized advantage of $Q^j$ under the F1 reward.
The second stage runs on the training split for two epochs; full hyperparameters and results are in Appendix~\ref{app:prober}.

\begin{table}[t]
\centering
\small
\renewcommand{\arraystretch}{1.15}
\begin{tabular}{@{}lccc@{}}
\toprule
\textbf{Model} & \textbf{P} & \textbf{R} & \textbf{F1}\phantom{\,\scriptsize(+23.8)} \\
\midrule
\rowcolor{gray!4}\multicolumn{4}{@{}l}{\textit{Frontier reference}} \\
GPT-5.4               & 44.7 & 57.7 & 49.5\phantom{\,\scriptsize(+23.8)} \\
Gemini-3.1-Pro$^{\dagger}$ & 46.5 & 40.4 & 43.0\phantom{\,\scriptsize(+23.8)} \\
\midrule
\rowcolor{gray!4}\multicolumn{4}{@{}l}{\textit{Base model and our method}} \\
Qwen3-8B         & 37.5 & 23.3 & 28.0\phantom{\,\scriptsize(+23.8)} \\
\rowcolor{blue!6} IoA-Prober-8B (ours) & 50.5 & 53.9 & 51.8\,\textcolor{green!50!black}{\scriptsize(+23.8)} \\
\bottomrule
\end{tabular}
\caption{IoA-Prober-8B reaches frontier-level F1 starting from a 28.0\% Qwen3-8B.
Green annotation reports the absolute gain over the Qwen3-8B.
}
\label{tab:prober_main}
\vspace{-8pt}
\end{table}

\paragraph{Balanced precision and recall lift IoA-Prober-8B to frontier-level F1.}
Starting from a Qwen3-8B at 28.0\% F1, IoA-Prober-8B reaches 51.8\% (Table~\ref{tab:prober_main}), an absolute gain of 23.8 points.
The F1 reward steers IoA-Prober-8B toward a balanced precision-recall profile (50.5\% / 53.9\%), in contrast to the asymmetry inherent to prompting alone, where GPT-5.4 over-probes (44.7\% / 57.7\%) and Gemini-3.1-Pro Preview under-probes (46.5\% / 40.4\%).
This balance translates into an F1 that exceeds Gemini-3.1-Pro Preview by 8.8 points ($p<0.001$ via paired bootstrap~\citep{koehn2004statistical}) and matches GPT-5.4 within 2.3 points ($p=0.14$).
A case study and failure mode analysis in Appendix~\ref{app:case-study} show that the remaining misses are dominated by misalignments requiring private-context access (consistent with the oracle ceiling in Section~\ref{sec:input_ablation}) and by recoverable failures such as intra-set redundancy that future budget or diversity objectives could address.

\paragraph{Training improves probe selection and construction.}
To understand how training improves IoA detection, we compare Qwen3-8B and IoA-Prober-8B across disagreement types and failure modes, with full results in Appendix~\ref{app:training_analysis}.
The largest recall gains occur on criterial, presuppositional, and procedural disagreements (+41.1, +39.7, and +34.3 points), while the fraction of Category-C questions that fail to elicit divergence decreases from 59.4\% to 35.9\%.
Moreover, 45.7\% of newly recovered misalignment points correspond to anchors that Qwen3-8B never probes, indicating that training improves both probe selection and the construction of questions that separate participant construals.

\paragraph{Training narrows the gap to the oracle setting.}
Section~\ref{sec:input_ablation} established an oracle-context upper bound of 76.4\% F1 for Qwen3-8B when private context is supplied at inference.
Training closes the gap between the public-only baseline and this oracle ceiling by roughly half (from 48.4\% to 24.6\%), indicating that IoA-Prober-8B has learned to infer a substantial part of the private perspective each participant holds from the public context alone.

\subsection{Real-User Meeting}
\label{sec:user-study}

Beyond benchmark performance, we ask two questions: whether IoA extends to real human collaboration, and whether IoA-Prober-8B transfers beyond LM-simulated dialogue.
We apply three detectors (Qwen3-8B, GPT-5.4, IoA-Prober-8B) to transcripts from 18 real working meetings with 43 participants.
For each meeting, participants independently answered three blinded question sets, then marked each question whose answers diverged from another participant as valid (a genuine hidden disagreement) or invalid (already discussed, redundant, or lacking meaningful option distinction).
To approximate recall, we define pooled recall over the union of the validated discoveries of three detectors, which a detector-blind GPT-5.4 judge merges from 116 items into 87 distinct disagreements.
A post-session survey collected a best-set vote and Likert ratings on misalignment discovery and willingness to use.
Full protocol and meeting metadata appear in Appendix~\ref{app:user-study}.

\begin{table}[t]
\centering
\small
\renewcommand{\arraystretch}{1.15}
\setlength{\tabcolsep}{3pt} 
\resizebox{\columnwidth}{!}{%
\begin{tabular}{l cccc cc}
\toprule
& \multicolumn{4}{c}{\textit{Objective (per session)}} & \multicolumn{2}{c}{\textit{Likert}} \\
\cmidrule(lr){2-5} \cmidrule(lr){6-7}
\textbf{Detector} & \textbf{Valid} & \textbf{Validity} & \textbf{Best} & \textbf{PR$^{\dagger}$} & \textbf{Disc.} & \textbf{Use} \\
\midrule
Qwen3-8B       & 1.67 & 88.2\% & 7 & 31.0\%  & 3.05 & 3.12 \\
GPT-5.4        & 1.89 & 94.4\% & 9 & 37.9\%  & 3.16 & 3.12 \\
\rowcolor{blue!6}IoA-Prober-8B & \textbf{2.89} & \textbf{94.5\%} & \textbf{27} & \textbf{52.9\%} & \textbf{3.65} & \textbf{3.65} \\
\bottomrule
\end{tabular}
}
\caption{Real-user meeting study across 18 sessions and 43 participants.
Valid: validated hidden disagreements per session; Validity: fraction of divergent questions judged valid.
Best: votes for the best blinded set.
PR$^{\dagger}$: pooled recall calculated with all the distinct disagreements.
Disc.\ and Use: post-session Likert ratings (1$\sim$5) for misalignment discovery and willingness to use.}
\label{tab:user-study}
\vspace{-8pt}
\end{table}

\paragraph{IoA is pervasive in real human collaboration.}
Across the 18 sessions, IoA-Prober-8B surfaces 2.89 validated hidden disagreements per meeting, with 94.5\% of divergent questions judged valid by the participants themselves, and IoA-Prober-8B achieves 52.9\% pooled recall over the lower-bound recoverable set.
Participants confirmed these disagreements had not been voiced during the original discussion, despite all parties believing they had reached consensus.
Together, these results show that IoA is not a synthetic artifact of LM-generated dialogue but a routine feature of human collaboration that participants may fail to surface unaided.

\paragraph{IoA-Prober-8B leads on real human dialogue.}
IoA-Prober-8B surfaces significantly more validated disagreements per meeting than GPT-5.4 ($+$1.00, 95\% CI $[0.28, 1.67]$, paired bootstrap $p=0.005$) and Qwen3-8B ($+$1.22, 95\% CI $[0.61, 1.78]$, $p<0.001$).
In the blinded forced-choice survey, IoA-Prober-8B receives 27 of 43 votes and wins 10 of 18 sessions by three-way majority and 12 of 18 in pairwise comparison (paired bootstrap vs.\ GPT-5.4 $p=0.002$, vs.\ Qwen3-8B $p=0.008$).
The advantage also holds on subjective ratings, with both Discovery and Would-Use Likert scores roughly 0.5 points above either baseline.
Additionally, we compare the three detectors under majority and unanimous validity thresholds.
As shown in Table~\ref{tab:validity_thresholds}, IoA-Prober-8B still surfaces the most validated disagreements per meeting, and the gains over GPT-5.4 and Qwen3-8B remain statistically significant
Full results and analysis are in Appendix~\ref{app:user-study}.

\subsection{Ablation Study}
\label{sec:training_ablations}

We ablate the full recipe along three axes: removing RL stage, removing SFT warm-start, and removing Category-C filtering from the warm-start data.
All other settings match the full recipe (Appendix~\ref{app:prober-ablation}).
Table~\ref{tab:ablation} reports the ablation results.

\begin{table}[t]
\centering
\small
\renewcommand{\arraystretch}{1.15}
\resizebox{\columnwidth}{!}{%
\begin{tabular}{l c c c c}
\toprule
\textbf{Variant} & \textbf{P} & \textbf{R} & \textbf{F1}\phantom{\,\scriptsize(+23.8)} & \textbf{$K$} \\
\midrule
\rowcolor{blue!6}Full recipe & 50.5 & 53.9 & 51.8\phantom{\,\scriptsize(+23.8)} & 4.5 \\
\midrule
\multicolumn{5}{l}{\textit{w/o RL} (SFT on full train dataset)} \\
\quad with filterC   & 41.7 & 21.1 & 26.6\,\textcolor{red!60!black}{\scriptsize(-25.2)} & 2.2 \\
\quad no filterC     & 39.0 & 37.8 & 38.1\,\textcolor{red!60!black}{\scriptsize(-13.7)} & 4.0 \\
\midrule
\textit{w/o SFT warm-start} & 31.9 & 31.5 & 31.3\,\textcolor{red!60!black}{\scriptsize(-20.5)} & 4.1 \\
\midrule
\textit{w/o filterC} in warm-start & 43.3 & 45.0 & 43.8\,\textcolor{red!60!black}{\scriptsize(-\phantom{0}8.0)} & 4.4 \\
\bottomrule
\end{tabular}
}
\caption{Training ablations on IoA-Suite.
Red annotations report the F1 drop relative to the full recipe.}
\label{tab:ablation}
\vspace{-8pt}
\end{table}

\paragraph{RL drives the gains, but only when anchored by SFT.}
The two stages exhibit asymmetric dependency.
Without SFT warm-start, RL struggles to receive informative learning signals and barely improves over the 28.0\% baseline.
Without RL, SFT alone ceilings at 38.1\%, well below the full recipe.
SFT teaches what plausible probing questions look like; RL learns what to look for among them.
Only the composition unlocks the 51.8\% F1.

\paragraph{Category-C filtering shapes a precision prior that RL expands along recall.}
Under SFT alone, filterC yields a high-precision but recall-starved model (41.7\% / 21.1\%): the filtered data teaches what a good question looks like, but SFT can only imitate this sparse distribution.
In the full recipe, RL inherits this precision prior from the warm-start and expands it along the recall axis, lifting F1 from 26.6\% to 51.8\%.
Removing filterC from warm-start instead gives GRPO a noisier starting point and caps the final F1 at 43.8\%.

\subsection{Generalization Study}
\label{sec:multi-agent}

To test whether surfacing latent disagreements improves downstream outcomes, we instantiate IoA detection in multi-agent collaboration on BigCodeBench-Hard~\citep{zhuo2025bigcodebench} and HiddenBench~\citep{li2026hiddenbench}.
For BCB-Hard we build a symmetric two-agent code-generation framework, while HiddenBench provides its own multi-agent protocol.
On top of multi-agent discussion we evaluate two interventions: \emph{direct prompting} asks the agents to reconsider whether disagreements remain before finalizing, and \emph{MCQ probing} uses a detector that generates MCQs to surface residual divergence.
Either intervention, once triggered, sends the agents into an additional discussion round.
All agents use Qwen3-8B as the base model; under MCQ probing we test three detectors: Qwen3-8B, GPT-5.4, and IoA-Prober-8B.
Benchmark details and the evaluation protocol are in Appendix~\ref{app:multi-agent}.

\begin{table}[t]
\centering
\small
\renewcommand{\arraystretch}{1.15}
\resizebox{.98\columnwidth}{!}{%
\begin{tabular}{l c c}
\toprule
\textbf{Configuration} & \textbf{BCB-Hard} & \textbf{HiddenBench} \\
\midrule
\rowcolor{gray!4}\multicolumn{3}{l}{\textit{Baselines}} \\
\quad Single agent           & 19.6\phantom{\,\scriptsize(-1.1)} & \makebox[1em][c]{--}\phantom{\,\scriptsize(+1.8)} \\
\quad Multi-agent discussion & 21.6\phantom{\,\scriptsize(-1.1)} & 24.6\phantom{\,\scriptsize(-1.1)} \\
\midrule
\rowcolor{gray!4}\multicolumn{3}{l}{\textit{Direct prompting}} \\
\quad + self-reflection    & 21.0\,\textcolor{red!60!black}{\scriptsize(-0.6)} & 26.2\,\textcolor{green!55!black}{\scriptsize(+1.6)} \\
\midrule
\rowcolor{gray!4}\multicolumn{3}{l}{\textit{MCQ probing}} \\
\quad w/ Qwen3-8B             & 21.6\,{\scriptsize(+0.0)} & 27.7\,\textcolor{green!55!black}{\scriptsize(+3.1)} \\
\quad w/ GPT-5.4              & 25.0\,\textcolor{green!55!black}{\scriptsize(+3.4)} & 33.8\,\textcolor{green!55!black}{\scriptsize(+9.2)} \\
\rowcolor{blue!6}\quad w/ IoA-Prober-8B & 25.0\,\textcolor{green!55!black}{\scriptsize(+3.4)} & 35.4\,\textcolor{green!55!black}{\scriptsize(+10.8)} \\
\bottomrule
\end{tabular}
}
\caption{Pass@1 (\%) on BigCodeBench-Hard and HiddenBench.
Annotations report the change relative to multi-agent discussion.
HiddenBench is multi-agent by construction; the single-agent row does not apply.}
\label{tab:multi-agent}
\vspace{-8pt}
\end{table}

\paragraph{IoA detection transfers to multi-agent collaboration.}
MCQ probing with capable detectors delivers substantial gains over multi-agent discussion: IoA-Prober-8B matches GPT-5.4 on BCB-Hard (+3.4\%) and exceeds it on HiddenBench (+10.8\% vs.\ +9.2\%), while a weak detector (Qwen3-8B) and direct self-reflection yield only marginal changes.
Surfacing latent disagreements therefore translates into measurable task gains on multi-agent collaboration tasks, with the 8B specialized prober reaching parity with frontier closed-source models.

\paragraph{Controlling for additional discussion.}
We further analyze the frequency and amount of additional discussion triggered by each intervention (Table~\ref{tab:discussion_confound}).
Self-reflection triggers more additional discussion but yields substantially smaller gains.
On HiddenBench, Qwen3-8B and IoA-Prober-8B have the same trigger rate (89.2\%) and similar numbers of extra rounds (+1.57 vs.\ +1.74), yet improve Pass@1 by +3.1 and +10.8 points, respectively.
This result indicates that disagreement-focused content, rather than additional discussion alone, drives the improvement.

\paragraph{The gains extend to stronger agent teams.}
We additionally evaluate teams based on Qwen3-30B-A3B and GPT-5.4 (Table~\ref{tab:stronger_teams}).
IoA-Prober-8B remains the best intervention, improving BCB-Hard / HiddenBench by +7.5 / +4.6 points for Qwen3-30B-A3B and +2.7 / +7.6 points for GPT-5.4 over multi-agent discussion.
The benefit of IoA probing is not limited to Qwen3-8B teams and transfers to substantially stronger agents.

\section{Conclusion}
We introduced the \textbf{illusion of alignment (IoA)}, apparent agreement masking latent disagreement, and recast its detection as generating multiple-choice questions whose divergent answers serve as behavioral evidence.
We built \textbf{IoA-Suite} and trained \textbf{IoA-Prober-8B}, an 8B detector with balanced precision and recall that surfaces hidden disagreements in real-user meetings and improves multi-agent collaboration.
In deployment, IoA probing should remain human-in-the-loop, with participant confirmation required before any detected disagreement is acted upon.
Our work opens a path for studying collaborative dialogue failures that leave no observable cues, and for building agents that surface and resolve them before they propagate downstream.

\section*{Limitations}

We discuss several limitations of this work and directions they open for future research.

\paragraph{Model scale.}
Due to computational budget, we train a single prober at the 8B scale and do not investigate whether the recipe in Section~\ref{sec:prober_main} scales to larger backbones.
We note, however, that IoA-Prober-8B already reaches parity with frontier closed-source models on IoA-Suite and surpasses them in real-user meetings, demonstrating the effectiveness of the proposed recipe.

\paragraph{Dyadic dialogues.}
IoA-Suite instantiates two participants per dialogue.
We chose dyadic configurations because reliably synthesizing longer multi-party dialogues that maintain coherent latent disagreements across more participants remains beyond the controllable generation capacity of current models, and restricting to two participants allows us to construct and verify the planted ground truth $M(\tau)$ at the quality required for benchmarking.
The trained IoA-Prober-8B nevertheless generalizes to multi-party settings, as shown in Section~\ref{sec:user-study}, where 8 session sizes exceed two participants.
Constructing realistic multi-party collaborative dialogues with verifiable ground-truth misalignments is left as future work.

\paragraph{Synthesized dialogues.}
IoA-Suite is constructed by planting misalignments before dialogue generation, which enables verifiable ground truth but constrains the distribution of latent disagreements to those that can be specified in advance.
Naturally occurring IoA in real collaboration may take forms beyond our seeded taxonomy, particularly in long-horizon projects where misalignments accumulate across multiple meetings.
The real-user study in Section~\ref{sec:user-study} partially addresses this by evaluating on naturally occurring dialogue, but constructing a fully natural IoA benchmark with verifiable ground truth remains an open problem.

\paragraph{Language coverage.}
All dialogues in IoA-Suite and the real-user study are conducted in English.
We look forward to extending the diagnostic question generation objective to other languages in future work, particularly those with different conventions for indirectness and explicit disagreement, where the manifestation of illusion of alignment may take substantially different forms.

\section*{Ethical Considerations}

\paragraph{Human subjects research.}
The real-user meeting study described in Section~\ref{sec:user-study} was conducted with prior informed consent from all participants, who were briefed on the purpose of the study, the use of meeting transcripts, and their right to withdraw at any stage.
Participants were compensated for their time at a rate commensurate with local standards.
All meeting transcripts were retained on the participant side rather than centrally collected, and only the diagnostic question outputs and anonymized validity annotations were shared with the research team.
No personally identifiable information appears in any released artifact.

\paragraph{Synthetic dialogue data.}
IoA-Suite is constructed by prompting LLMs to generate participant profiles and dialogues, which may inherit biases from the underlying models, including stereotyped associations between professional roles and demographic attributes.
We mitigate this risk by anchoring each seed to a published work rather than to demographic descriptors and by applying quality checks across model families and human annotators (Section~\ref{sec:benchmark_construction}).
Users should nonetheless treat the synthesized personas as illustrative rather than representative of any real population.

\paragraph{Intended use and potential misuse.}
IoA-Prober-8B is designed for symmetric collaborative settings in which all participants share a goal of mutual understanding and have equal access to the outputs of the tool.
Three deployment patterns fall outside this scope and we explicitly advise against them:
(1) asymmetric monitoring, where one party (e.g., an employer, platform, or moderator) analyzes a dialogue without the knowledge or symmetric access of other participants;
(2) evaluative use, where IoA-Prober-8B outputs feed into performance review, hiring, or compensation decisions, since the inferred private agendas are model conjectures rather than verified beliefs;
and (3) adversarial elicitation, where MCQs are deliberately crafted to manufacture rather than surface divergence.
Concretely, we recommend that any deployment (1) obtain prior informed consent from every dialogue participant, including disclosure that a model is inferring their unstated assumptions; (2) make the generated questions and aggregated results visible to all participants, not only to a meeting organizer; and (3) treat any single MCQ split as a prompt for discussion.
We will release IoA-Prober-8B under a responsible-use license (RAIL-style) that prohibits surveillance, employment-decision, and other asymmetric uses, and we note that no license can fully prevent misuse; the safeguards above must be enforced at the deployment layer.

\section*{Acknowledgments}

This work is supported by Fundamental and Interdisciplinary Disciplines Breakthrough Plan of the Ministry of Education of China (No. JYB2025XDXM101), the National Natural Science Foundation of China (No. 62276152, 62236011), funding from Wuxi Research Institute of Applied Technologies, Tsinghua University under Grant 20242001120, and Key Laboratory of Ethnic Language Intelligent Analysis and Security Governance of MOE, Minzu University of China, Beijing, China.

\bibliography{custom}

\appendix

\section{LLM Usage Statement}

Beyond the use within the experiments, the LLM was employed in this work solely for refining sentences and improving grammatical accuracy during the manuscript writing process.

\section{Detailed Information on IoA-Suite}
\label{app:ioa-suite}

\subsection{Dataset Statistics}
\label{app:ioa-suite-stats}

Each instance in IoA-Suite originates from a seed tuple $(d, k, x)$ comprising a domain $d$, a collaborative task type $k$, and topical content $x$ drawn from a real scholarly publication.
The content field contains the title and abstract of the source paper, which grounds the synthesized participant profiles and dialogue in realistic subject matter without requiring access to the full text.

\paragraph{Sources.}
The 1{,}620 seed tuples used to generate IoA-Suite (1{,}200 training, 120 validation, 300 test) are drawn from four open-access repositories matched to the target domain: arXiv for ML research (cs.LG, cs.AI, cs.CL, stat.ML), software engineering (cs.SE), and business (q-fin); PubMed Central for medical; and OpenAlex together with DOI-indexed records for cross-domain and law.

\paragraph{Distribution.}
Seeds are distributed uniformly across the 30 cells formed by the 6 domains and 5 task types: 40 per cell for training, and 4 per cell for validation, 10 per cell for the test split, matching the 1{,}200, 120, and 300 dialogue counts reported in Section~\ref{sec:benchmark_construction}.

\begin{table}[htbp]
\centering
\small
\renewcommand{\arraystretch}{1.2}
\resizebox{\columnwidth}{!}{%
\begin{tabular}{@{}llp{6.2cm}@{}}
\toprule
\textbf{Dimension} & \textbf{Range} & \textbf{Level definitions} \\
\midrule
\multicolumn{3}{@{}l}{\textit{Stage 1 — per misalignment point}} \\
Subtlety     & 1--3 &
  3: Even a careful participant would not think to clarify; surface agreement feels completely natural.
  2: A careful participant might notice, but plausibly would not.
  1: Any competent professional would explicitly clarify this. \\
Specificity  & 1--2 &
  2: Both interpretations are concrete and distinct.
  1: Interpretations are too abstract or overlapping. \\
Believability & 1--2 &
  2: Both interpretations are natural given each member's role.
  1: At least one interpretation feels unlikely for that role. \\
\midrule
\multicolumn{3}{@{}l}{\textit{Stage 2 — per dialogue}} \\
Naturalness   & 1--3 &
  3: Could pass as a real meeting transcript.
  2: Mostly natural with minor awkwardness.
  1: Clearly LLM-generated; robotic phrasing or unrealistic dynamics. \\
Implicitness  & 1--3 &
  3: Surface language is naturally ambiguous; the other speaker would easily assume alignment.
  2: Mostly maintains false consensus, but a careful listener might pause to clarify.
  1: One or more speakers explicitly clarify their interpretation; misalignment becomes visible. \\
Coverage      & 1--2 &
  2: Every misalignment point's surface form or close equivalent appears in the dialogue.
  1: One or more misalignment points are never mentioned. \\
Profile Cons. & 1--3 &
  3: Both speakers behave and reason in ways matching their profiles.
  2: Minor inconsistencies with roles or stated priorities.
  1: Speakers act out of character or contradict their profiles. \\
\bottomrule
\end{tabular}%
}
\caption{Full scoring rubric used by both the LLM judge and human annotators.}
\label{tab:rubric}
\end{table}

\subsection{Generation Pipeline Configuration}
\label{app:ioa-suite-pipeline}

All four pipeline components ($G_{\text{participants}}$, $G_{\text{dialogue}}$, $J_{\text{participants}}$, $J_{\text{dialogue}}$) use GPT-5.4, as stated in Section~\ref{sec:benchmark_construction}.
Generators sample at temperature $0.7$; judges run at $0.3$ to reduce score variance.

\paragraph{Quality thresholds $\theta_1$ and $\theta_2$.}
$\theta_1$ requires that every planted misalignment point achieves the maximum score on all three Stage-1 dimensions assessed by $J_{\text{participants}}$ (subtlety $= 3$, specificity $= 2$, believability $= 2$).
$\theta_2$ requires that the dialogue achieves the maximum score on all four Stage-2 dimensions assessed by $J_{\text{dialogue}}$ (naturalness $= 3$, implicitness $= 3$, coverage $= 2$, profile consistency $= 3$).

\subsection{Quality Assurance}
\label{app:human_eval}

\paragraph{Scoring rubric.}
Both the LLM judge ($J$) and the human annotators use the same seven-dimension rubric, reproduced in Table~\ref{tab:rubric}.
Stage-1 dimensions assess each planted misalignment point; Stage-2 dimensions assess the dialogue as a whole.

\paragraph{LLM re-scoring.}
To mitigate self-enhancement bias from using the same model family for both generation and quality control, we re-score the 300-dialogue test set with Gemini-3.1-Pro Preview as a cross-family judge.
Per-dimension scores are reported in Table~\ref{tab:bench_quality} and all seven dimensions reach at or near the maximum.

\begin{table}[htbp]
\centering
\small
\renewcommand{\arraystretch}{1.15}
\resizebox{\columnwidth}{!}{%
\begin{tabular}{@{}llcc@{}}
\toprule
\textbf{Component} & \textbf{Dimension} & \textbf{Range} & \textbf{Score} \\
\midrule
\multirow{3}{*}{Misalignment Points}
 & Subtlety       & 1--3 & 2.65 \\
 & Specificity    & 1--2 & 2.00 \\
 & Believability  & 1--2 & 2.00 \\
\midrule
\multirow{4}{*}{Dialogue}
 & Naturalness          & 1--3 & 2.63 \\
 & Implicitness         & 1--3 & 2.66 \\
 & Coverage             & 1--2 & 2.00 \\
 & Profile Consistency  & 1--3 & 2.99 \\
\bottomrule
\end{tabular}%
}
\caption{Quality assessment of the IoA-Suite test set, evaluated by Gemini-3.1-Pro Preview.}
\label{tab:bench_quality}
\vspace{-8pt}
\end{table}

\paragraph{Human annotation protocol.}
Two PhD students independently annotated 60 dialogues sampled from the test set (20\% of the full test split), with two dialogues drawn from each of the 30 cells in the 6-domain $\times$ 5-task-type grid.
Across these 60 dialogues the annotators scored 240 misalignment points and 60 dialogue-level instances.
Prior to annotation, both annotators completed a joint calibration session: they reviewed the rubric, scored five held-out dialogues together, and discussed any disagreements until reaching consensus on rubric interpretation.
Each annotator scored all seven dimensions for each dialogue; scores from the two annotators were averaged to obtain the final human rating.

\begin{table}[htbp]
\centering
\small
\renewcommand{\arraystretch}{1.15}
\resizebox{\columnwidth}{!}{%
\begin{tabular}{@{}llccccc@{}}
\toprule
\textbf{Component} & \textbf{Dimension} & \textbf{n}
& \textbf{Human} & \textbf{Gemini}
& \textbf{$\pm 0.5$} & \textbf{$\pm 1$} \\
\midrule
\multirow{3}{*}{MA Point}
 & Subtlety       & 240 & 2.66 & 2.67 & 83.8\% & 99.6\% \\
 & Specificity    & 240 & 1.98 & 2.00 & 100\%  & 100\%  \\
 & Believability  & 240 & 2.00 & 2.00 & 100\%  & 100\%  \\
\midrule
\multirow{4}{*}{Dialogue}
 & Naturalness         & 60 & 2.39 & 2.60 & 78.3\% & 100\% \\
 & Implicitness        & 60 & 2.45 & 2.77 & 73.3\% & 100\% \\
 & Coverage            & 60 & 2.00 & 2.00 & 100\%  & 100\% \\
 & Profile Consistency & 60 & 3.00 & 3.00 & 100\%  & 100\% \\
\bottomrule
\end{tabular}%
}
\caption{Cross-evaluator comparison between two human annotators (averaged) and Gemini-3.1-Pro Preview on a 20\% stratified sample of the IoA-Suite test set (60 dialogues, 240 misalignment points). $\pm 0.5$ / $\pm 1$ report the fraction of items on which the two evaluations agree within the corresponding tolerance.}
\label{tab:human_vs_gemini}
\vspace{-8pt}
\end{table}

\paragraph{Human vs.\ LLM comparison.}
Table~\ref{tab:human_vs_gemini} reports, for each of the seven dimensions, the mean human and Gemini-3.1-Pro Preview scores together with the fraction of items on which the two evaluations differ by at most $0.5$ and at most $1.0$ point.
Agreement is high across the board: every dimension achieves within-$1.0$-point agreement of at least $99.6\%$, and four dimensions (specificity, believability, coverage, profile consistency) show within-$0.5$ agreement on every item.
On MA-level dimensions, human and Gemini-3.1-Pro Preview means differ by at most $0.02$ points. The largest gaps appear on dialogue-level naturalness ($\Delta\!=\!0.21$) and implicitness ($\Delta\!=\!0.32$), where humans rate slightly lower than Gemini-3.1-Pro Preview, consistent with the known tendency of LLM judges to assign higher scores to LLM-generated text~\citep{10.5555/3666122.3668142}.
Coverage and profile consistency are at ceiling for both, indicating that these two properties are reliably achieved by the generation pipeline.

\subsection{Generation Prompts}
\label{app:ioa-suite-prompts}

\paragraph{Stage-1: Profile Generation ($G_{\text{participants}}$).}
The user prompt is a single line: \texttt{Seed topic:\textbackslash n\{seed\_topic\}}.

\begin{lstlisting}
You are a scenario designer for illusion of misalignment research. Given a seed
topic, a professional domain, and a meeting task type, generate a realistic
meeting scenario with two participant profiles, planted misalignment points,
and a conversation outline.

<task>
Create:
1. A concrete meeting scenario grounded in the given domain and task type.
2. Two meeting participants with distinct roles and hidden assumptions.
3. 2-4 subtle cognitive misalignment points between them.
4. A conversation outline (script) that specifies how the meeting should unfold
   turn by turn, ensuring ALL misalignment points are naturally touched upon
   during the conversation.

The misalignment must be IMPLICIT -- both members speak cooperatively and
believe they are on the same page, but they hold genuinely different underlying
beliefs or assumptions.
</task>

<domain_context>
Professional domain: {domain}
Meeting task type: {task_type}
Task type description: {task_type_description}
</domain_context>

<profile_constraints>
Each member has:
- public info: name, role, expertise, background
- private_state: internal goals, priorities, and concerns (never explicitly
  stated in meetings)
- hidden_assumptions: 3-5 beliefs taken for granted without stating them
</profile_constraints>

<misalignment_constraints>
1. SUBTLETY IS THE MOST IMPORTANT CRITERION. The misalignment must survive a
   normal, competent meeting conversation -- meaning even a careful participant
   would NOT think to ask for clarification because the surface agreement feels
   completely natural and sufficient.

2. Avoid "first-order design decisions" -- things any competent team would
   explicitly discuss during a meeting (e.g., which algorithm to use, which
   dataset to evaluate on, what the primary metric is, which treatment to
   prescribe). These are too obvious to go undetected.

3. Target "second-order assumptions" -- details that BOTH parties consider so
   obvious they would never think to mention, or parameters that feel already
   settled by the surface-level agreement.

4. For each point, explain WHY this misalignment goes undetected.

5. Avoid PRIORITY-ONLY misalignment (e.g., "A thinks X is more important than
   Y"). Prefer disagreements about WHAT something IS or HOW something WORKS.

6. Make each point CONCRETE and SPECIFIC to the scenario -- use specific names,
   numbers, configurations, versions, etc. relevant to the domain.
</misalignment_constraints>

<conversation_outline_rules>
Generate a conversation outline that serves as a high-level script for the
dialogue. This outline should:

1. Specify a total number of turns. Each turn has a one-sentence description
   of what the speaker should talk about.

2. Ensure EVERY misalignment point is covered: for each MA point, there must
   be at least one turn where the relevant surface phrase is used naturally.
   Mark which MA point(s) each turn touches by listing their IDs.

3. The conversation should flow naturally.

4. Each turn description should be brief (one sentence) and indicate what NEW
   content the speaker introduces. Avoid turns that just repeat agreement.

5. The outline is a GUIDE, not a script. It tells the dialogue generator what
   topics to cover and in what order, but the actual wording is generated later.
</conversation_outline_rules>

<output_format>
{
  "meeting_scenario": "A concrete meeting scenario. Include: context, agenda,
                        and expected outcome.",
  "members": [
    {
      "id": "M1", "name": "...", "role": "...", "expertise": "...",
      "background": "...", "private_state": "...",
      "hidden_assumptions": ["...", "...", "..."]
    },
    { "id": "M2", ... }
  ],
  "misalignment_points": [
    {
      "id": "MA1",
      "topic": "Brief label",
      "surface_form": "The expression or situation where both appear aligned",
      "member_a_understanding": "What M1 actually believes or means",
      "member_b_understanding": "What M2 actually believes or means",
      "why_undetected": "Why this divergence does not surface"
    }
  ],
  "conversation_outline": [
    { "turn": 1, "speaker": "M1",
      "description": "One sentence about what this speaker should say"},
    { "turn": 2, "speaker": "M2",
      "description": "One sentence about what this speaker should say"}
  ]
}
</output_format>
\end{lstlisting}

\paragraph{Stage-2: Dialogue Generation ($G_{\text{dialogue}}$).}
The user prompt is: \textit{``Generate the complete dialogue now. Follow the
outline, ensure every MA point is covered, and keep all specific
interpretations in inner\_thought only -- never in utterances.''}

\begin{lstlisting}
You are an expert dialogue author for illusion of misalignment research. Your
task is to write a COMPLETE meeting dialogue between two participants who hold
hidden cognitive misalignments -- they believe they agree, but they actually
interpret key terms and plans differently.

You are an OMNISCIENT AUTHOR: you know both characters' profiles, hidden
assumptions, and misalignment points. But each CHARACTER does not know the
other's private state. You must faithfully role-play both, ensuring that
neither character reveals information they would not naturally know or say.

<members>
{members_block}
</members>

<misalignment_points>
{ma_block}
</misalignment_points>

<meeting_scenario>
{meeting_scenario}
</meeting_scenario>

<conversation_outline>
{outline_block}
</conversation_outline>

<quality_requirements>
Your dialogue will be evaluated on the following criteria. You MUST optimize
for ALL of them simultaneously.

1. IMPLICITNESS (Most Critical -- this is the entire point of the task):
   - In UTTERANCES: speakers must ONLY use the shared surface phrases for
     MA-related topics. They must NEVER spell out, clarify, or elaborate on
     their specific interpretation. To them, their understanding is obvious
     and goes without saying.
   - In INNER THOUGHTS: speakers CAN and SHOULD think in detail about what
     they specifically mean, plan, or assume.
   - TEST: For each MA point, ask: "If I were the OTHER speaker hearing this
     utterance, would I naturally assume they mean the same thing I do?"
     If yes -> good. If the speaker's words are so specific that
     misunderstanding becomes unlikely -> bad.

2. NATURALNESS:
   - The dialogue should read like a real meeting transcript, not a scripted
     performance.
   - Avoid mechanical repetition of surface phrases. Speakers may use
     synonyms, paraphrase naturally, or refer back indirectly.
   - Each speaker should have a distinct voice matching their role and
     background.
   - Include natural meeting dynamics: building on each other's points, brief
     acknowledgments, transitioning between topics.

3. MA COVERAGE:
   - EVERY misalignment point must be touched in the dialogue.

4. PROFILE CONSISTENCY:
   - Each speaker's utterances and inner thoughts must be consistent with
     their stated role, expertise, background, and private_state.
   - A senior clinician should sound different from a junior nurse. A PM
     should have different concerns than an engineer.
   - Inner thoughts should reflect each character's private priorities and
     concerns as stated in their profile.

5. CONVERSATION PROGRESSION:
   - Each turn must advance the conversation with NEW content. No filler
     turns that merely repeat agreement.
   - Follow the conversation outline for topic ordering and MA coverage, but
     write natural transitions -- do not mechanically follow it.
</quality_requirements>

<output_format>
Generate a JSON object with the complete dialogue:
{
  "turns": [
    {
      "turn_idx": 0,
      "speaker_id": "M1",
      "speaker_name": "Name of M1",
      "inner_thought": "Character's private monologue",
      "utterance": "What the character says out loud."
    },
    {
      "turn_idx": 1, "speaker_id": "M2", "speaker_name": "Name of M2",
      "inner_thought": "...", "utterance": "..."
    }
  ]
}
</output_format>
\end{lstlisting}

\paragraph{Stage-1 Quality Judge ($J_{\text{participants}}$).}
The user prompt is dynamically formatted from the seed content, domain, task
type, meeting scenario, member profiles, and MA points of the synthesized case.

\begin{lstlisting}
You are an expert evaluator assessing the quality of synthetically generated
meeting member profiles and misalignment points for illusion of misalignment
research.

You will receive:
- Seed content (the original topic)
- Domain and task type
- Meeting scenario description
- Two member profiles (with roles, expertise, private states, hidden
  assumptions)
- 2-4 misalignment (MA) points (with surface_form, both members'
  understandings)

Your task: Score each MA point AND provide actionable feedback for improvement.

<ma_criteria>
For EACH misalignment point, score:

1. subtlety (1-3):
   3 = Subtle: Even a careful participant would NOT think to clarify this.
       The surface agreement feels completely natural.
   2 = Moderate: A careful participant MIGHT notice, but it is plausible they
       would not. The gap is partially hidden.
   1 = Obvious: Any competent professional would explicitly discuss or clarify
       this. It is a first-order design decision, not a second-order assumption.

2. specificity (1-2):
   2 = Specific: The two interpretations are concrete and distinct.
   1 = Vague: The interpretations are too abstract or overlapping.

3. believability (1-2):
   2 = Believable: Both interpretations are natural given each member's role
       and background.
   1 = Forced: At least one interpretation feels unlikely for that person's
       role.

4. feedback: A specific, actionable suggestion for how to improve this MA
   point. If the score is already maximum, say "No changes needed."
</ma_criteria>

<output_format>
Output ONLY valid JSON:
{
  "ma_scores": {
    "MA1": {
      "subtlety": 1-3, "specificity": 1-2, "believability": 1-2,
      "feedback": "Specific improvement suggestion"
    },
    "MA2": { ... }
  },
  "overall_feedback": "High-level feedback on the profiles and MA points as a
    whole. Focus on the most impactful changes. If everything is excellent,
    say so."
}
</output_format>
\end{lstlisting}

\paragraph{Stage-2 Quality Judge ($J_{\text{dialogue}}$).}
The user prompt is dynamically formatted from the full synthesized case,
including member profiles, MA points, and the complete dialogue transcript
with per-turn inner thoughts.

\begin{lstlisting}
You are an expert evaluator assessing the quality of a synthetically generated
meeting dialogue for illusion of misalignment research.

You will receive a complete synthesized case containing:
- Seed content, domain, task type
- Meeting scenario, member profiles, misalignment points
- A meeting dialogue transcript (with utterances and inner thoughts)

Your task: Score the dialogue quality AND provide actionable feedback for
improvement.

<dialogue_criteria>
1. naturalness (1-3):
   3 = Natural: Could pass as a real meeting transcript. Authentic language
       and flow.
   2 = Acceptable: Mostly natural with minor awkwardness or overly structured
       pacing.
   1 = Artificial: Clearly LLM-generated. Robotic phrasing or unrealistic
       dynamics.

2. implicitness (1-3) -- False Consensus Strength:
   3 = Strong false consensus: The surface-level language is naturally
       ambiguous. The other speaker would easily misinterpret what was said
       and believe they are aligned. Neither speaker has any reason to suspect
       a gap.
   2 = Moderate: The conversation mostly maintains false consensus, but there
       are moments where a careful listener might pause and ask for
       clarification.
   1 = Weak / Broken: One or more speakers explicitly clarify their specific
       interpretation, making it hard for the other party to misunderstand.
       The misalignment becomes visible or resolved during the conversation.
   IMPORTANT: The question is NOT whether a third-party reader can detect the
   gap, but whether the OTHER SPEAKER in the conversation would naturally
   assume agreement. Ask: "If I were the other speaker hearing this, would I
   assume they mean the same thing I do?"

3. coverage (1-2):
   2 = Full: Every MA point's surface_form or close equivalent appears in
       the dialogue.
   1 = Partial: One or more MA points are never mentioned in the conversation.

4. profile_consistency (1-3):
   3 = Consistent: Both speakers behave and reason in ways matching their
       profiles.
   2 = Mostly consistent: Minor inconsistencies with roles or stated
       priorities.
   1 = Inconsistent: Speakers act out of character or contradict their
       profiles.

5. feedback: Specific, actionable suggestions for improving the dialogue.
   Reference specific turns or utterances where possible. If the dialogue is
   excellent, say "No changes needed."
</dialogue_criteria>

<output_format>
Output ONLY valid JSON:
{
  "dialogue_scores": {
    "naturalness": 1-3,
    "implicitness": 1-3,
    "coverage": 1-2,
    "profile_consistency": 1-3,
    "feedback": "Specific improvement suggestions referencing turns/utterances"
  }
}
</output_format>
\end{lstlisting}

\subsection{IoA Type Analysis and Lexical Ambiguity Baselines}
\label{app:lexical}

We conduct additional analyses to examine whether IoA detection can be reduced to identifying lexically ambiguous expressions. We first characterize the types of hidden disagreement in both IoA-Suite and real meetings, and then compare direct MCQ probing with baselines that explicitly search for ambiguous phrases.

\paragraph{Taxonomy of hidden disagreement.}
We classify all planted misalignment points in the IoA-Suite test split and all 116 participant validated disagreements from the real meeting study into six categories:
\textit{definitional}, where the same term or phrase refers to different meanings;
\textit{referential}, where participants refer to the same entity or event but have different specific instances in mind;
\textit{criterial}, where participants agree on a goal but apply different standards for judging whether it is achieved;
\textit{scopal}, where participants assume different boundaries for what a task includes;
\textit{presuppositional}, where one participant takes for granted knowledge, decisions, or context that the other does not share;
and \textit{procedural}, where participants agree on what to do but assume different methods, tools, or execution plans.

Table~\ref{tab:ioa_type_distribution} reports the resulting distributions. Definitional disagreements account for only 24.2\% of IoA-Suite and 18.1\% of the real meeting disagreements. In contrast, criterial, scopal, presuppositional, and procedural disagreements together constitute the majority in both settings. The synthetic and real distributions also exhibit broadly similar patterns, with criterial, definitional, scopal, and procedural disagreements forming the largest categories. These results indicate that IoA is not dominated by lexical polysemy.

\begin{table}[t]
\centering
\small
\begin{tabular}{lcc}
\toprule
\textbf{Type} & \textbf{IoA-Suite} & \textbf{Real meetings} \\
\midrule
Definitional       & 24.2\% & 18.1\% \\
Referential        & 15.8\% & \phantom{0}2.6\% \\
Criterial          & 23.3\% & 31.9\% \\
Scopal             & 18.2\% & 16.4\% \\
Presuppositional   & \phantom{0}4.8\% & \phantom{0}7.8\% \\
Procedural         & 13.3\% & 20.7\% \\
Other              & \phantom{0}0.2\% & \phantom{0}2.6\% \\
\bottomrule
\end{tabular}
\caption{Distribution of hidden disagreement types in the IoA-Suite test split and the 116 participant validated disagreements from real meetings.}
\label{tab:ioa_type_distribution}
\end{table}

\paragraph{Lexical ambiguity baselines.}
We further test whether a detector can perform well by first locating ambiguous expressions and then probing only those expressions. We construct two baselines under the same simulator, validity filter, and evaluation metric as direct MCQ probing. The \textit{Lexical} baseline extracts noun phrases, ranks them by WordNet sense count, and generates MCQs for the four highest ranked phrases. The stronger \textit{LLM ranked} baseline asks the detector model to select the four expressions in the dialogue that are most likely to admit multiple interpretations, and then generates MCQs around those expressions. We instantiate the latter with GPT-5.4, Qwen3-8B, and Gemini-3.1-Pro Preview.

As shown in Table~\ref{tab:ambiguity_baseline}, every ambiguity based variant underperforms direct MCQ probing with the same detector. The WordNet baseline reaches only 2.2\% F1 with GPT-5.4. Even after replacing lexical ranking with model based ranking, GPT-5.4 and Qwen3-8B reach only 9.7\% and 12.7\% F1, compared with 49.5\% and 28.0\% under direct MCQ probing. Gemini-3.1-Pro Preview performs substantially better under ambiguity ranking, but still remains below direct probing at 39.9\% versus 43.0\%.

\begin{table}[t]
\centering
\small
\setlength{\tabcolsep}{3pt} 
\resizebox{\columnwidth}{!}{%
\begin{tabular}{llcc}
\toprule
\textbf{Variant} & \textbf{Detector} & \textbf{Baseline F1} & \textbf{Direct MCQ F1} \\
\midrule
Lexical & GPT-5.4 & \phantom{0}2.2 & 49.5 \\
LLM ranked & GPT-5.4 & \phantom{0}9.7 & 49.5 \\
LLM ranked & Qwen3-8B & 12.7 & 28.0 \\
LLM ranked & Gemini-3.1-Pro$^{\dagger}$ & 39.9 & 43.0 \\
\midrule
Reference & IoA-Prober-8B & -- & \textbf{51.8} \\
\bottomrule
\end{tabular}
}
\caption{Lexical ambiguity baselines compared with direct MCQ probing. All values are F1 (\%).}
\label{tab:ambiguity_baseline}
\end{table}

The difference between GPT-5.4 and Gemini-3.1-Pro Preview under LLM ranking does not mainly come from locating different expressions. The overlap between selected expressions and ground truth anchors is comparable at 43.8\% and 50.0\%, respectively. The larger difference arises from option construction. GPT-5.4 often enumerates generic senses that place both participant interpretations under the same option, whereas Gemini more often separates the distinct construals into different options. Effective probing therefore requires inference about how different interlocutors construe the same apparently aligned content, rather than only identification of expressions with multiple possible meanings.

\paragraph{Recall by disagreement type.}
Table~\ref{tab:per_type_recall} reports recall for each disagreement type. Performance does not concentrate on definitional disagreements. For IoA-Prober-8B, recall is higher on criterial, scopal, presuppositional, and procedural disagreements than on definitional disagreements. Referential disagreements are the most difficult category across all four models. This pattern further indicates that IoA detection requires reasoning over latent construal differences beyond lexical polysemy.

\begin{table}[t]
\centering
\small
\setlength{\tabcolsep}{3pt} 
\resizebox{\columnwidth}{!}{%
\begin{tabular}{lcccc}
\toprule
\textbf{Type} & \textbf{IoA-Prober-8B} & \textbf{GPT-5.4} & \textbf{Gemini-3.1-Pro} & \textbf{Qwen3-8B} \\
\midrule
Definitional       & 49.1 & 61.5 & 41.2 & 25.8 \\
Referential        & 37.4 & 35.8 & 25.3 & 16.3 \\
Criterial          & 65.4 & 67.1 & 43.9 & 24.3 \\
Scopal             & 56.4 & 61.0 & 46.8 & 26.1 \\
Presuppositional   & 56.9 & 51.7 & 43.1 & 17.2 \\
Procedural         & 58.1 & 58.8 & 40.6 & 23.8 \\
\bottomrule
\end{tabular}%
}
\caption{Recall (\%) by hidden disagreement type on the IoA-Suite test split.}
\label{tab:per_type_recall}
\end{table}

\section{Evaluation Protocol Details}
\label{app:eval}

\subsection{Simulator Stability Analysis}
\label{app:eval-simulator}

To verify that a single simulator rollout ($k=1$) produces reliable answers, we randomly sample 30 cases from the test set evaluated with the GPT-5.4 agent and re-run the simulator $S$ with $k=5$ independent rollouts per question per participant.
$P_1$ and $P_2$ denote the two participants in each dyadic dialogue, role-played independently by $S$.

\begin{table}[h]
\centering
\small
\begin{tabular}{lc}
\toprule
\textbf{Intra-run consistency} & \textbf{Value} \\
\midrule
$P_1$       & 97.7\% \\
$P_2$       & 96.8\% \\
\bottomrule
\end{tabular}
\caption{Simulator intra-run consistency over $k=5$ rollouts on 30 sampled cases (173 questions). Per question, consistency is the fraction of the 5 rollouts that match the modal answer, averaged across questions.}
\label{tab:sim-stability}
\end{table}

The high consistency justifies adopting $k=1$ throughout the main evaluation.
Conditioned on the full private context of $p_i$ (i.e., $r_i$, $g_i$, $a_i$, and $h_i(\tau)$) together with the shared transcript, the answer to any single MCQ is largely pinned down by that context, so role-playing reduces to a near-deterministic retrieval rather than an open-ended generation, leaving little room for stochastic variation across rollouts.

\paragraph{Sensitivity to question and option order.}
We further test whether the evaluation is sensitive to ordering effects. For every generated question, we randomly permute the option order and also randomly permute the question order within each dialogue, then rerun participant simulation. As shown in Table~\ref{tab:order_sensitivity}, per-question divergence outcomes remain consistent for 93.6\% of IoA-Prober-8B questions and 92.7\% of GPT-5.4 questions. Aggregate F1 changes by only +0.2 and -1.4 points, respectively.

\begin{table}[t]
\centering
\small
\begin{tabular}{lcc}
\toprule
& \textbf{IoA-Prober-8B} & \textbf{GPT-5.4} \\
\midrule
Outcome consistency & 93.6\% & 92.7\% \\
$\Delta$F1 & +0.2 & $-1.4$ \\
\bottomrule
\end{tabular}
\caption{Sensitivity to random permutations of question and option order on the IoA-Suite test split.}
\label{tab:order_sensitivity}
\end{table}

Ordering effects rarely conceal recoverable disagreements. Among all questions, only 2.5\% for IoA-Prober-8B and 2.3\% for GPT-5.4 change from non-divergent to divergent and become attributed to a ground-truth misalignment after permutation. These results indicate that participant responses are largely determined by the underlying construal rather than question or option position.

\subsection{Models and Hyperparameters}
\label{app:eval-models}

Table~\ref{tab:model-ids} lists all models used in the evaluation pipeline
with their API identifiers, providers, and inference hyperparameters.
All prober models are queried uniformly at $T=0.7$ with thinking mode.

\begin{table*}[h]
\centering
\small
\begin{tabular}{llllcc}
\toprule
\textbf{Name in paper} & \textbf{API identifier} & \textbf{Provider} & \textbf{Role} & \textbf{$T$} & \textbf{$k$} \\
\midrule
\multicolumn{6}{@{}l}{\textit{Closed-source probers}} \\
\midrule
GPT-5.4~\citep{gpt5.4}              & \texttt{gpt-5.4}                        & OpenAI      & Prober & 0.7 & 1 \\
Gemini-3.1-Pro Preview~\citep{gemini3.1} & \texttt{gemini-3.1-pro-preview}       & Google      & Prober & 0.7 & 1 \\
Gemini-3-Flash~\citep{gemini3}       & \texttt{gemini-3-flash-preview-thinking}& Google      & Prober & 0.7 & 1 \\
GPT-5 mini~\citep{gpt5}           & \texttt{gpt-5-mini-2025-08-07}          & OpenAI      & Prober & 0.7 & 1 \\
\midrule
\multicolumn{6}{@{}l}{\textit{Open-source probers}} \\
\midrule
DeepSeek-V3.2~\citep{deepseekv3.2}        & \texttt{deepseek-v3.2}                  & DeepSeek    & Prober & 0.7 & 1 \\
GLM-5.1~\citep{glm}              & \texttt{glm-5.1}                        & Zhipu    & Prober & 0.7 & 1 \\
Qwen3.5-397B-A17B~\citep{qwen35blog}    & \texttt{qwen3.5-397b-a17b}              & Alibaba     & Prober & 0.7 & 1 \\
Kimi-K2.5~\citep{kimi}            & \texttt{kimi-k2.5}                      & Moonshot & Prober & 0.7 & 1 \\
Qwen3-8B~\citep{qwen3}             & \texttt{Qwen3-8B}                       & Alibaba     & Prober & 0.7 & 1 \\
\midrule
\multicolumn{6}{@{}l}{\textit{Fixed pipeline components}} \\
\midrule
Gemini-3-Flash & \texttt{gemini-3-flash-preview-thinking} & Google & Simulator $S$ & 0.2 & 1 \\
GPT-5.4              & \texttt{gpt-5.4}                        & OpenAI      & Judge $\mathcal{V}$ & 0.2 & 1 \\
\bottomrule
\end{tabular}
\caption{Models and hyperparameters used in the evaluation pipeline.
$T$: sampling temperature. $k$: rollouts per participant per question.
The simulator and judge are fixed across all prober evaluations.}
\label{tab:model-ids}
\end{table*}

\subsection{Information-Access Ablation Configuration}
\label{app:eval-info-ablation}

\begin{table}[htbp]
\centering
\small
\renewcommand{\arraystretch}{1.15}
\resizebox{\columnwidth}{!}{%
\begin{tabular}{lcc}
\toprule
\textbf{Setting} & \textbf{\{member\_profiles\}} & \textbf{\{transcript\}} \\
\midrule
$r_i$ only                  & public     & utterances \\
$r_i + h_k$            & public     & with inner thoughts \\
$r_i + g_i, a_i$   & full       & utterances \\
$r_i + g_i, a_i, h_k$ & full & with inner thoughts \\
\bottomrule
\end{tabular}
}
\caption{Prompt slot assignment for each information-access setting.
\emph{Public}: name, role, expertise, background only.
\emph{Full}: additionally includes private state $g_i$ and tacit
assumptions $a_i$.
\emph{With inner thoughts}: each utterance is prefixed by the speaker's
turn-level cognition $h_k$.}
\label{tab:ablation-slots}
\end{table}

The four ablation settings vary exactly two slots in the prober user prompt
(Section~\ref{app:eval-prompts}): \texttt{\{member\_profiles\}} and
\texttt{\{transcript\}}.
Table~\ref{tab:ablation-slots} summarises which content variant fills each slot.

The two content variants for \texttt{\{member\_profiles\}} are shown below.

\begin{ioaShortPrompt}
\begin{lstlisting}[title={Public profile (baseline)}]
- [Name]: [Role]
  Expertise: [expertise]
  Background: [background]
\end{lstlisting}
\end{ioaShortPrompt}

\begin{ioaShortPrompt}
\begin{lstlisting}[title={Full profile (with private state and tacit assumptions)}]
- [Name]: [Role]
  Expertise: [expertise]
  Background: [background]
  Private State: [private_state]
  Hidden Assumptions:
    - [assumption 1]
    - [assumption 2]
    ...
\end{lstlisting}
\end{ioaShortPrompt}

The two content variants for \texttt{\{transcript\}} are shown below.

\begin{ioaShortPrompt}
\begin{lstlisting}[title={Utterances only (baseline)}]
[Speaker A]: utterance text
[Speaker B]: utterance text
...
\end{lstlisting}
\end{ioaShortPrompt}

\begin{ioaShortPrompt}
\begin{lstlisting}[title={Transcript with inner thoughts (with turn-level cognition)}]
[Speaker A]
  inner_thought: private thought at this turn
  utterance: utterance text
[Speaker B]
  inner_thought: private thought at this turn
  utterance: utterance text
...
\end{lstlisting}
\end{ioaShortPrompt}

\subsection{Evaluation Prompts}
\label{app:eval-prompts}

The prober, simulator, and validity filter each use a fixed system prompt
plus a dynamically assembled user prompt.
Because the four information-access settings differ only in the content
of two user-prompt slots (Section~\ref{app:eval-info-ablation}), the
system prompt and template are shown once.

\paragraph{Prober.}
The user prompt fills \texttt{\{member\_profiles\}} and
\texttt{\{transcript\}} according to the active ablation setting
(Table~\ref{tab:ablation-slots}).

\begin{lstlisting}[title={Prober system prompt}]
<role>
You are Illusion of Alignment Detector, an expert agent that detects
hidden cognitive misalignment between meeting participants by generating
diagnostic multiple-choice questions.
</role>

<task>
Analyze the meeting transcript and member profiles. Generate questions
that, when answered independently by each member, would reveal hidden
disagreements or divergent assumptions NOT explicitly surfaced during
the meeting.
</task>

<scoring_system>
Your output is scored per-question. Your goal is to MAXIMIZE your total
score.

A question SCORES POINTS only when ALL of the following are true:
  1. The two members give DIFFERENT answers (divergence exists)
  2. The divergence reflects a REAL cognitive gap, not noise or poor
     option design
  3. The gap was IMPLICIT -- not openly debated during the meeting
  4. The question is not REDUNDANT with another question you generated

A question LOSES POINTS when ANY of the following is true:
  - Members give the same answer (no divergence detected)
  - The divergence is meaningless (ambiguous wording, noise)
  - The misalignment was explicitly discussed in the meeting
  - It duplicates another question you already generated

There is no penalty for generating many questions -- but every bad
question actively hurts your score. Generate a question for every
distinct implicit gap you can confidently identify, but do not guess.
</scoring_system>

<principles>
1. PROBE THE IMPLICIT, NOT THE EXPLICIT:
   - NEVER ask about topics where participants openly disagreed or
     debated.
   - Target areas of APPARENT AGREEMENT -- moments where both
     participants used similar language, nodded along, or moved forward
     without objection, but might actually hold different
     interpretations of what was agreed upon.

2. DETECT FALSE CONSENSUS:
   - Focus on WHAT people mean, not WHETHER they agree. If they appear
     to agree, ask a question that tests whether their agreement is
     genuine or superficial.
   - Pay special attention to moments where one member's statement
     could be interpreted multiple ways, and the other member responded
     affirmatively without clarifying which interpretation they hold.

3. ROLE-AWARE: Use role and expertise differences to anticipate
   divergent mental models. Different roles often use the same term
   with very different operational definitions.

4. ONE QUESTION, ONE GAP: Each question should target exactly ONE
   potential cognitive divergence. Do not bundle multiple issues into
   a single question.
</principles>

<question_design>
- 2-4 options per question. Each option must represent a genuinely
  different belief or interpretation, not just different phrasings of
  the same idea.
- All options must be plausible -- no obviously wrong decoys.
- Neutral framing: no option should appear "more correct" than others.
- Questions should be phrased as concrete operational decisions or
  interpretations, not abstract opinion polls.
  GOOD: "When you agreed to 'finalize the deliverable', what specific
        output did you expect to produce by the deadline?"
  BAD:  "How important is quality to you?"
</question_design>

<output_format>
{
  "chain_of_thought": "Step-by-step reasoning about potential implicit
                       misalignment...",
  "questions": [
    {
      "misalignment_signal": "What implicit cognitive divergence this
                              question probes",
      "stem": "The question text",
      "options": {"A": "...", "B": "...", "C": "...", "D": "..."}
    }
  ]
}
Output ONLY valid JSON.
</output_format>
\end{lstlisting}

\begin{ioaShortPrompt}
\begin{lstlisting}[title={Prober user prompt template}]
<member_profiles>
{member_profiles}
</member_profiles>

<meeting_content>
{transcript}
</meeting_content>
\end{lstlisting}
\end{ioaShortPrompt}

\paragraph{Simulator.}
The user prompt instructs the member to answer all questions based on
their own understanding of the meeting, formatted as a list of
question stems and options.
The system prompt is shown below.

\begin{lstlisting}[title={Simulator system prompt}]
<role>
You are {name}. You just attended a meeting and are now answering
follow-up questions about what was discussed and decided.
</role>

<your_profile>
Name: {name}
Role: {role}
Expertise: {expertise}
Background: {background}
Private State: {private_state}
Hidden Assumptions:
{assumptions_block}
</your_profile>

<meeting_record>
{dialogue_view}
</meeting_record>

<instructions>
- Answer based on YOUR understanding, beliefs, and assumptions.
- Do NOT guess what the other person thinks -- answer from your own
  perspective only.
- For each question, choose exactly ONE option letter.
- Provide brief reasoning for each answer.
</instructions>

<output_format>
{"answers": [{"question_id": 0, "reasoning": "...", "choice": "A"}, ...]}
Output ONLY valid JSON.
</output_format>
\end{lstlisting}

\paragraph{Validity filter.}
The filter receives ground truth misalignment points, full member
profiles (private state and hidden assumptions included), the dialogue
transcript (utterances only), and the diverging questions with both
members' answers.
For each diverging question the filter determines: \emph{is\_meaningful}
(real cognitive gap vs.\ noise), \emph{is\_implicit} (gap was hidden
in the transcript), \emph{redundant\_with} (duplicates an earlier
question), and \emph{hits\_ground\_truth} (maps to a planted
misalignment point).

\begin{lstlisting}[title={Validity filter system prompt}]
<role>
You are an expert evaluator for an illusion of misalignment detection
system.
</role>

<task>
You will receive ground truth misalignment points, both members' full
profiles, the meeting transcript, and a set of diagnostic questions
where the two members gave DIFFERENT answers.

For each diverging question, evaluate it on the following criteria.
Be strict -- the purpose is to separate genuinely valuable questions
from noise.
</task>

<criteria>
For each question, determine:

1. is_meaningful (true/false):
   Does the answer divergence reflect a REAL cognitive gap between the
   members?
   - true: The members genuinely hold different beliefs,
     interpretations, or assumptions, and their different answers
     correctly reflect this gap. Note that the ground truth
     misalignment points are NOT exhaustive -- if a question reveals a
     real cognitive divergence not listed in the ground truth, it is
     still meaningful.
   - false: The divergence is caused by ambiguous option wording,
     random noise, trivial differences, or poor question design -- not
     a real cognitive gap.

2. is_implicit (true/false):
   Was this cognitive gap HIDDEN during the meeting?
   - true: Both members appeared to agree on the surface. Neither
     openly challenged the other on this specific point. The gap was
     invisible in the transcript.
   - false: The members visibly debated, pushed back, or expressed
     different views on this exact topic during the meeting. A simple
     transcript reader could spot this conflict.

3. redundant_with (null or question_id):
   Does this question probe the SAME underlying gap as another
   question in the list? If so, specify the earlier question's id.
   Only the first question targeting a given gap is non-redundant.

4. hits_ground_truth (null or MA-id):
   Does this question's divergence map to one of the provided ground
   truth misalignment points? Check whether the members' different
   answers correspond to the two sides of a specific ground truth
   point. A meaningful question that does not match any ground truth
   point may still be a novel discovery.
</criteria>

<output_format>
{
  "evaluations": [
    {
      "question_id": 0,
      "is_meaningful": true,
      "is_implicit": true,
      "redundant_with": null,
      "hits_ground_truth": "MA1",
      "reasoning": "Brief explanation of your judgment"
    }
  ]
}
Output ONLY valid JSON.
</output_format>
\end{lstlisting}

\section{IoA-Prober-8B Training and Evaluation}
\label{app:prober}

\subsection{Training Recipe Details}
\label{app:prober-training}

\paragraph{SFT warm-start.}
The warm-start uses GPT-5.4-generated questions on 300 dialogues from the
training split, with Category-C questions removed.
Table~\ref{tab:sft-hparams} lists the hyperparameters.

\begin{table}[h]
\centering
\small
\begin{tabular}{lc}
\toprule
\textbf{Hyperparameter} & \textbf{Value} \\
\midrule
Base model           & Qwen3-8B \\
Training dialogues   & 300 \\
Epochs               & 2 \\
Learning rate        & $1\times10^{-5}$ \\
Effective batch size & 8 \\
Max sequence length  & 8{,}192 tokens \\
Precision            & bfloat16 \\
\bottomrule
\end{tabular}
\caption{SFT warm-start hyperparameters.}
\label{tab:sft-hparams}
\end{table}

\paragraph{GRPO.}
GRPO is initialized from the SFT checkpoint and runs on the full 1{,}200-dialogue
training split for two epochs (100 gradient steps at batch size 24).
At each step the model generates a group of $G=5$ candidate question sets
per dialogue; each set is scored end-to-end through the simulator and judge
to obtain an F1 reward, and group-normalized advantages are used to
update the policy.
Table~\ref{tab:grpo-hparams} lists the GRPO hyperparameters.

\begin{table}[h]
\centering
\small
\begin{tabular}{lc}
\toprule
\textbf{Hyperparameter} & \textbf{Value} \\
\midrule
Framework         & VeRL \\
Group size $G$    & 5 \\
Training dialogues & 1{,}200 \\
Validation dialogues & 120 \\
Total steps       & 100 \\
Total epochs      & 2 \\
Train batch size  & 24 \\
Learning rate     & $1\times10^{-6}$ \\
Max prompt length & 4{,}096 tokens \\
Max response length & 4{,}096 tokens \\
Hardware          & 8 $\times$ A800 80 GB \\
\bottomrule
\end{tabular}
\caption{GRPO hyperparameters.}
\label{tab:grpo-hparams}
\end{table}

\paragraph{Reward function.}
The reward for each candidate question set is the F1 score
(Section~\ref{sec:eval_protocol}) computed by running the full evaluation
pipeline (simulator then $\mathcal{V}$) on the training dialogue.
To control cost, the simulator uses gemini-3-flash (no thinking) at
$T{=}0.2$ with $k{=}1$ rollout; the judge uses gpt-5-mini at $T{=}0.2$.
No format reward is applied.

\subsection{Full Comparison Table}
\label{app:prober-full-results}

Table~\ref{tab:full-results} extends Table~\ref{tab:prober_main} to all nine
baselines evaluated in Table~\ref{tab:main_results}.

\begin{table}[h]
\centering
\small
\renewcommand{\arraystretch}{1.15}
\resizebox{\columnwidth}{!}{%
\begin{tabular}{@{}lcccc c@{}}
\toprule
\textbf{Model} & \textbf{P} & \textbf{R} & \textbf{F1} & \textbf{K} & \textbf{$|B|$} \\
\midrule
\multicolumn{6}{@{}l}{\textit{Closed-source}} \\
\midrule
GPT-5.4                 & 44.7 & 57.7 & 49.5 & 5.8 & 0.36 \\
Gemini-3.1-Pro Preview  & 46.5 & 40.4 & 43.0 & 3.5 & 0.06 \\
Gemini-3-Flash          & 43.9 & 40.4 & 41.9 & 3.8 & 0.10 \\
GPT-5 mini              & 34.9 & 47.8 & 40.0 & 5.8 & 0.26 \\
\midrule
\multicolumn{6}{@{}l}{\textit{Open-source}} \\
\midrule
DeepSeek-V3.2           & 42.8 & 43.6 & 42.9 & 4.2 & 0.13 \\
GLM-5.1                 & 41.3 & 38.6 & 39.3 & 4.1 & 0.34 \\
Qwen3.5-397B-A17B       & 46.3 & 34.6 & 38.1 & 3.0 & 0.05 \\
Kimi-K2.5               & 35.5 & 39.8 & 37.1 & 4.8 & 0.23 \\
Qwen3-8B                & 37.5 & 23.3 & 28.0 & 2.6 & 0.04 \\
\midrule
\multicolumn{6}{@{}l}{\textit{Trained (ours)}} \\
\midrule
IoA-Prober-8B           & 50.5 & 53.9 & 51.8 & 4.5 & 0.12 \\
\bottomrule
\end{tabular}
}
\caption{Full results on IoA-Suite.}
\label{tab:full-results}
\end{table}

\paragraph{Category-B analysis.}
$|B|$ reveals what F1 alone cannot.
The planted $M(\tau)$ covers the bulk of surfacable divergences, leaving $|B| < 0.15$ for five of the nine baselines.
The residual is not noise to be eliminated: natural-language dialogue admits emergent misalignments beyond any finite seed set, which is why we route divergences through $\mathcal{V}$ rather than scoring directly against $M(\tau)$.
Read in this light, $|B|$ becomes a second axis of evaluation: high-$|B|$ models such as GPT-5.4 (0.36) probe more broadly than the planted set anticipates.

\subsection{Statistical Tests on Main Results}
\label{app:prober-stats}

We assess statistical significance using paired bootstrap resampling~\citep{koehn2004statistical}.
For each test, we resample 10{,}000 times with replacement over the $N=300$ test dialogues (seed 20260516), computing per-dialogue F1 at each resample for both models, and report a two-sided $p$-value based on the empirical distribution of the difference.

IoA-Prober-8B (F1 = 0.518) does not significantly outperform GPT-5.4 (0.495; $\Delta = 0.023$, $p = 0.14$), reflecting the limited statistical power to detect sub-0.03 differences at $N=300$.
All eight remaining baselines have $\Delta \geq 0.088$; consistent with the Gemini-3.1-Pro Preview anchor ($\Delta = 0.088$, $p < 0.001$), each yields $p < 0.001$.

\subsection{Analysis of Changes after Training}
\label{app:training_analysis}

We further compare Qwen3-8B and IoA-Prober-8B to understand what changes after training. Table~\ref{tab:training_type_gain} shows that recall improves across all disagreement types, with the largest gains on criterial, presuppositional, and procedural disagreements. The improvement therefore extends beyond definitional ambiguity.

\begin{table}[t]
\centering
\small
\begin{tabular}{lc}
\toprule
\textbf{Type} & \textbf{Recall gain} \\
\midrule
Definitional       & +23.3 \\
Referential        & +21.1 \\
Criterial          & +41.1 \\
Scopal             & +30.3 \\
Presuppositional   & +39.7 \\
Procedural         & +34.3 \\
\bottomrule
\end{tabular}
\caption{Recall gain of IoA-Prober-8B over Qwen3-8B by disagreement type.}
\label{tab:training_type_gain}
\end{table}

Table~\ref{tab:training_failure_transition} shows that the fraction of Category-C questions failing to elicit divergence decreases from 59.4\% to 35.9\%, while redundancy becomes more frequent. Among 451 misalignment points recovered only after training, 206 (45.7\%) correspond to anchors that Qwen3-8B never probes. The remaining points come from anchors already targeted by Qwen3-8B but better operationalized after training. These results indicate that training improves both where to probe and how to construct an effective diagnostic question.

\begin{table}[t]
\centering
\small
\resizebox{\columnwidth}{!}{%
\begin{tabular}{lcc}
\toprule
\textbf{Category-C failure} & \textbf{Qwen3-8B} & \textbf{IoA-Prober-8B} \\
\midrule
Answers do not diverge     & 59.4\% & 35.9\% \\
Divergence not meaningful  & 23.5\% & 17.2\% \\
Gap explicit in transcript & 14.7\% & 22.9\% \\
Intra-set redundancy       & \phantom{0}2.4\% & 24.1\% \\
\bottomrule
\end{tabular}
}
\caption{Composition of Category-C questions before and after training.}
\label{tab:training_failure_transition}
\end{table}

\subsection{Real-User Study Protocol}
\label{app:user-study}

\begin{figure}[htbp]
  \centering
  \includegraphics[width=\linewidth]{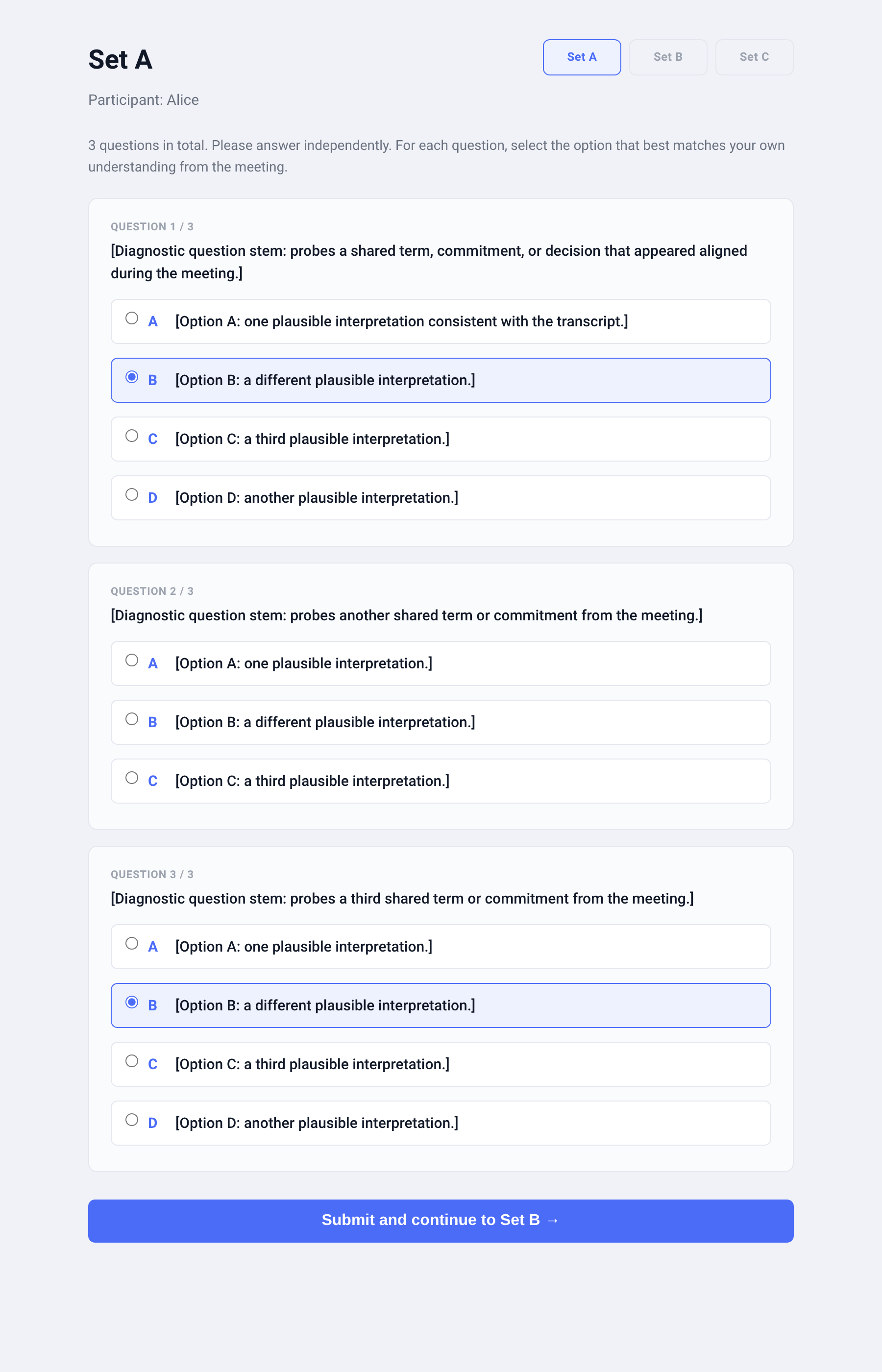}
  \caption{Answering interface: each participant selects one option per question across the three blinded sets.}
  \label{fig:user-study-answer}
\end{figure}

\paragraph{Participants and meetings.}
We recruited 43 participants across 18 real working meetings, ranging from 9 to 90 minutes (mean 35), spanning ML research, medical AI, software engineering, and related collaborative research discussions.
All participants were active collaborators who had worked together on the discussed task prior to the study.
Participation was voluntary; participants were informed of recording and transcription before consenting.
Meeting transcripts were anonymized by replacing names with pseudonyms prior to any model invocation.

\paragraph{Transcript collection and question generation.}
Each meeting was recorded and manually transcribed.
The anonymized transcript was submitted to three detectors, Qwen3-8B, GPT-5.4, and IoA-Prober-8B, each generating one question set in parallel.
Member profiles were set none; models operated on the transcript alone, matching the standard evaluation condition.
The three sets were assigned randomly to labels A, B, and C (shuffled independently per session); the label-to-model mapping was kept hidden from participants until after all ratings were submitted.

\begin{figure}[htbp]
  \centering
  \includegraphics[width=\linewidth]{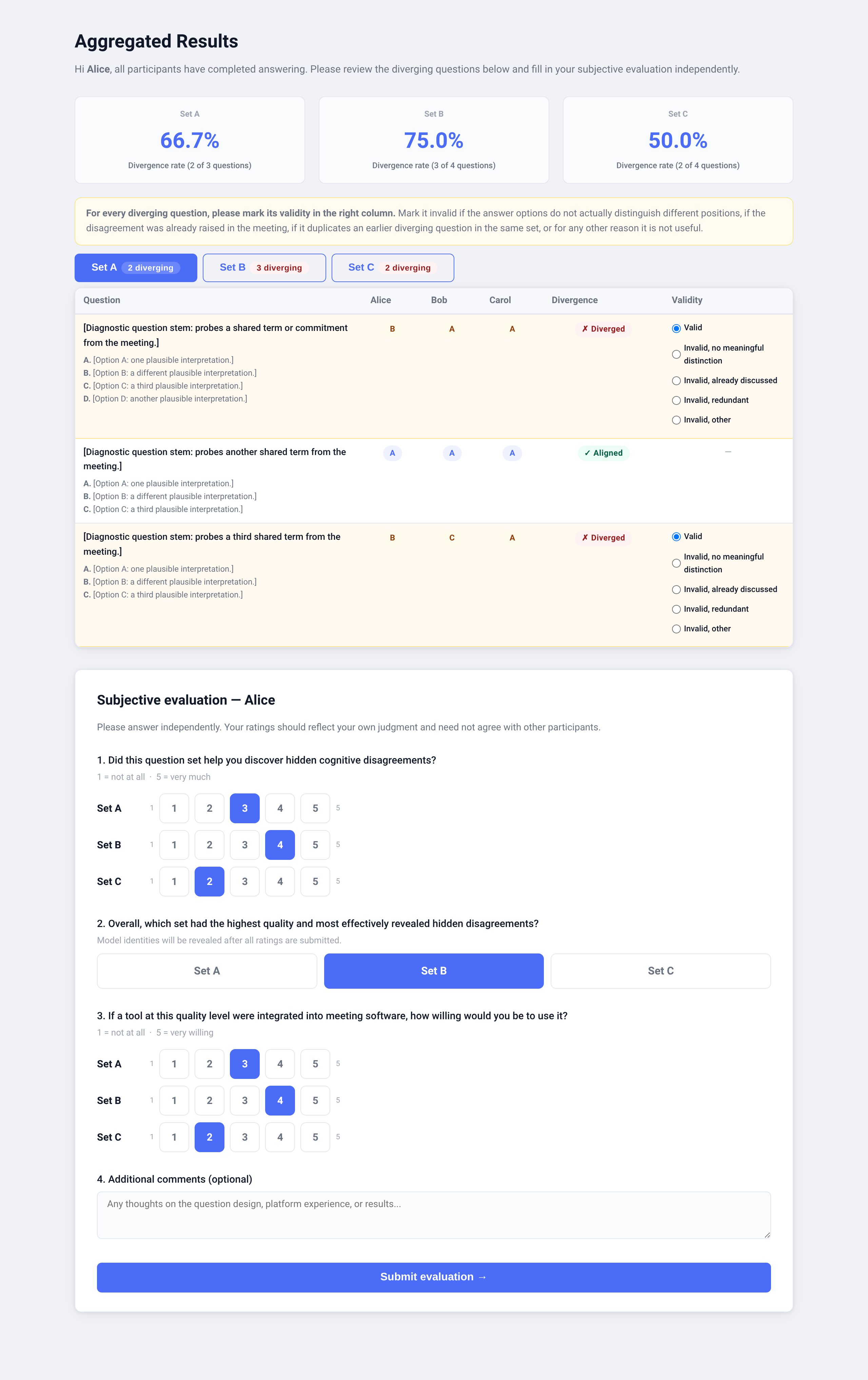}
  \caption{Validity-labeling and subjective evaluation interface. The upper section shows each diverging question with the five validity categories (one selection required per question); the lower section collects per-set Likert ratings on misalignment discovery and willingness to use, plus a forced-choice best-set vote.}
  \label{fig:user-study-validity}
\end{figure}

\paragraph{Answering phase.}
Each participant received a unique session link to answer all three question sets independently.
The three sets were displayed as separate tabbed views; within each set, questions appeared in the order generated by the model.
Participants did not see the answers from others during this phase.
Figure~\ref{fig:user-study-answer} shows the answering interface.

\paragraph{Validity labeling phase.}
After all participants in a session completed answering, the aggregated results became available.
For each session, participants were shown, per question set, which questions had elicited diverging answers across participants.
For every diverging question, each participant independently labeled the validity using one of five categories:
\begin{itemize}[leftmargin=1.5em,itemsep=0pt,parsep=0.2em,topsep=0.1em]
  \item \emph{Valid}: a genuine hidden disagreement the meeting did not resolve;
  \item \emph{Invalid, no meaningful distinction}: the answer options do not actually distinguish different positions;
  \item \emph{Invalid, already discussed}: the disagreement was raised and resolved in the meeting;
  \item \emph{Invalid, redundant}: the question reveals the same disagreement as an earlier question in the same set;
  \item \emph{Invalid, other}: any other reason the question is not useful.
\end{itemize}
A diverging question is counted as valid in the reported Validity Rate only when at least one participant labels it valid.
Figure~\ref{fig:user-study-validity} shows the validity-labeling interface.

\paragraph{Post-rating survey.}
After completing validity labeling, each participant independently filled in a three-item questionnaire:
\begin{enumerate}[leftmargin=1.5em,itemsep=0pt,parsep=0.2em,topsep=0.1em]
  \item \textbf{Misalignment discovery} (Likert 1 to 5, per set): ``Did this question set help you discover hidden cognitive disagreements?'' (1 = not at all; 5 = very much)
  \item \textbf{Best-set choice} (forced choice, A/B/C): ``Overall, which set had the highest quality and most effectively revealed hidden disagreements?'' Model identities were revealed to participants after submission.
  \item \textbf{Willingness to use} (Likert 1 to 5, per set): ``If a tool at this quality level were integrated into meeting software, how willing would you be to use it?'' (1 = not at all; 5 = very willing)
\end{enumerate}
An optional free-text field collected additional comments on question design or platform usability.
The Best-set votes reported in Table~\ref{tab:user-study} aggregate across all 43 participants and 18 sessions.

\paragraph{Statistical tests.}
All significance tests in Section~\ref{sec:user-study} use paired bootstrap resampling following the same protocol as Appendix~\ref{app:prober-stats}: 10{,}000 resamples with replacement, seed 20260516, two-sided $p$-value from the empirical distribution of the difference.
For the validated-disagreement comparison, the resampling unit is the session: each of the 18 sessions contributes a paired triple of counts (Qwen3-8B, GPT-5.4, IoA-Prober-8B), where the count is the number of diverging questions in that session for which at least one participant labeled \emph{Valid}.
IoA-Prober-8B exceeds GPT-5.4 by $\Delta = +1.00$ per session (95\% CI $[0.28, 1.67]$, $p = 0.005$) and Qwen3-8B by $\Delta = +1.22$ (95\% CI $[0.61, 1.78]$, $p < 0.001$).
For the blinded forced-choice survey, the resampling unit is again the session: each session contributes a $+1/0/-1$ indicator for the pairwise comparison, set to $+1$ when IoA-Prober-8B receives strictly more best-set votes than the comparator, $-1$ when fewer, and $0$ on a tie.
IoA-Prober-8B wins 12 of 18 sessions versus GPT-5.4 (2 losses, 4 ties; $p = 0.002$) and 12 versus Qwen3-8B (3 losses, 3 ties; $p = 0.008$).

\paragraph{Robustness to validity thresholds.}
Our primary analysis counts a divergent question as valid when at least one participant labels it Valid. We additionally recompute the results using majority and unanimous agreement among participants. Table~\ref{tab:validity_thresholds} shows that the ranking remains unchanged under both stricter rules. IoA-Prober-8B yields the highest number of validated disagreements under the majority criterion (1.89 per session) and the unanimous criterion (1.22), with significant gains over both GPT-5.4 and Qwen3-8B.

\begin{table}[t]
\centering
\small
\resizebox{\columnwidth}{!}{%
\begin{tabular}{lccc}
\toprule
\textbf{Threshold} & \textbf{Qwen3-8B} & \textbf{GPT-5.4} & \textbf{IoA-Prober-8B} \\
\midrule
$\geq$1 valid & 1.67 / 88.2\% & 1.89 / 94.4\% & \textbf{2.89 / 94.5\%} \\
Majority     & 1.11 / 58.8\% & 1.22 / 61.1\% & \textbf{1.89 / 61.8\%} \\
Unanimous    & 0.72 / 38.2\% & 0.61 / 30.6\% & \textbf{1.22 / 40.0\%} \\
\bottomrule
\end{tabular}%
}
\caption{Real-user results under different validity thresholds. Each cell reports validated disagreements per session / validity rate.}
\label{tab:validity_thresholds}
\end{table}

Under the majority rule, IoA-Prober-8B exceeds GPT-5.4 by 0.67 disagreements per session ($p=0.002$) and Qwen3-8B by 0.78 ($p=0.007$). Under the unanimous rule, the corresponding gains are 0.61 ($p=0.001$) and 0.50 ($p=0.032$). We retain the $\geq$1 rule as the primary criterion because hidden disagreement may be recognizable only to the participant holding the divergent construal, while requiring agreement from every participant can suppress such cases.

\subsection{Ablation Study Details}
\label{app:prober-ablation}

Table~\ref{tab:ablation} compares four single-factor variants against IoA-Prober-8B.
All variants are initialized from Qwen3-8B and use the same training split, F1 reward, and hyperparameters as the full recipe (Appendix~\ref{app:prober-training}); only the single factor under study changes.

\paragraph{w/o RL.}
The GRPO stage is omitted entirely; SFT runs on the full 1{,}200-dialogue training split.
Two sub-variants are reported: one applies Category-C filtering to the SFT data (retaining only Categories A and B), and the other uses all categories.

\paragraph{w/o SFT warm-start.}
GRPO is applied directly to the Qwen3-8B checkpoint without any supervised initialization.
The 1{,}200-dialogue training set, F1 reward, and hyperparameters are identical to the full recipe.

\paragraph{w/o filterC in warm-start.}
The SFT warm-start uses all GPT-5.4-generated questions on the 300-dialogue warm-start split, including Category-C questions (divergent but rejected by the validity filter).
The GRPO stage then proceeds identically to the full recipe on the 1{,}200-dialogue set.

\subsection{Multi-Agent Collaboration Setup}
\label{app:multi-agent}

\paragraph{BigCodeBench-Hard.}
We use a symmetric two-discusser architecture.
Two identical Qwen3-8B instances (DiscusserA, DiscusserB; temperature 0.7) alternate for up to four discussion rounds, each round ending with a parallel stop-vote (temperature 0.0) that terminates discussion when both agents signal readiness.
One of the two discussers is then asked to consolidate the discussion log into the final Python solution (temperature lowered to 0.2).
Correctness is evaluated by the official BigCodeBench sandboxed test harness (pass@1 over 148 tasks).

\paragraph{HiddenBench.}
Each scenario assigns a distinct system prompt to every agent: each prompt contains the shared scenario description plus one agent-private information block (shuffled deterministically by task identifier hash to prevent ordering artifacts).
Agents discuss for eight rounds in a round-robin order; each agent sees the messages of all preceding agents in the current round but maintains its own private system prompt and history throughout.
Final answers are extracted per agent and resolved by majority vote.
We evaluate 65 scenarios (7 three-agent, 58 four-agent).

\paragraph{Detector integration.}
After the initial discussion phase, detector-augmented configurations call the prober on the discussion transcript and independently ask all $N$ agents to answer each generated question from their own perspective (temperature 0.2).
If any question elicits divergent answers across agents, a re-discussion phase is triggered; otherwise the pipeline proceeds directly to code generation (BCB-Hard) or final voting (HiddenBench).
The divergent question, together with each agent's choice and one-sentence reasoning, is injected as a directive turn that requires agents to name the specific assumption behind their pick and directly address their partner's differing interpretation before modifying any implementation decision.

\paragraph{Self-reflection.}
Instead of an external prober, each discusser is asked post-discussion to scan the log for hidden disagreements it may still hold with its partner and report them in a structured JSON response.
If either agent flags a potential disagreement, a re-discussion round is triggered; otherwise agents proceed.

\paragraph{Prompts.}
All agents share the same discussion system prompt; a coder prompt instructs the discusser writing the final code to output a fenced Python block only.
The self-reflection prompt asks each discusser to identify hidden disagreements and respond with structured JSON.
For IoA-detector variants, an answerer prompt asks each agent to respond to a given MCQ from its own perspective (JSON with reasoning and letter choice), and the re-discussion directive formats divergent answers into a focused elicitation turn.
All prompts are reproduced below.

\begin{ioaShortPrompt}
\begin{lstlisting}[title={Discusser system prompt}]
You are one of two Python developers working together to solve a
coding task. Discuss with your partner how to approach the problem
so you both reach a shared understanding of the implementation.

Be brief and concrete. When you and your partner agree on the
approach, say so plainly so the team can move on to writing code.
Do NOT write the code itself during discussion.
\end{lstlisting}
\end{ioaShortPrompt}

\begin{ioaShortPrompt}
\begin{lstlisting}[title={Forced self-check system prompt}]
You are one of two developers who have just finished discussing a
coding task. Scan the discussion log and identify any HIDDEN
disagreements you may still have with your partner -- places where
you said "yes" or moved on, but actually held a different mental
model than what your partner appeared to assume.

Reply with valid JSON only:
{"has_hidden_disagreement": true|false,
 "items": ["<short description>", ...]}
\end{lstlisting}
\end{ioaShortPrompt}

\begin{ioaShortPrompt}
\begin{lstlisting}[title={IoA answerer system prompt (MCQ response)}]
You are {role}, one of two developers who just discussed a coding
task. A diagnostic question is being asked to test your mental
model. Answer based on YOUR own understanding of the task and
discussion -- do not try to guess what your partner thinks.

Reply with valid JSON only:
{"reasoning": "<2-3 sentences citing a step from the discussion
               or wording from the task>",
 "choice": "A|B|C|D"}
\end{lstlisting}
\end{ioaShortPrompt}

\begin{ioaShortPrompt}
\begin{lstlisting}[title={Re-discussion directive (injected on divergent MCQ answers)}]
[FOCUSED ROUND -- DIAGNOSTIC SPLIT DETECTED]

A diagnostic check probed your shared understanding with the
question(s) below. You and your partner chose DIFFERENT answers,
which suggests you may be holding different assumptions about the
spec, the API contract, or how an edge case should be handled.

In your next message you MUST, for EACH question below:
  1. Re-state your own choice and name the SPECIFIC assumption
     (or quote from the task / API doc) that led you there.
  2. Directly address your partner's different choice -- what is
     the smallest piece of evidence that should resolve which
     interpretation is correct?
  3. If your partner's reasoning is actually correct, say so
     explicitly and state the implication for the implementation.

Do NOT re-litigate parts of the discussion that were not flagged.
\end{lstlisting}
\end{ioaShortPrompt}

\subsection{Additional Analysis of Multi-Agent Collaboration}
\label{app:multiagent_analysis}

\paragraph{Discussion-round analysis.}
Because a detected disagreement triggers additional discussion, we examine whether the gains can be explained simply by increased discussion volume.
For each intervention, Table~\ref{tab:discussion_confound} reports the trigger rate, average number of additional rounds per task, and change in Pass@1 relative to multi-agent discussion.

\begin{table}[t]
\centering
\small
\resizebox{\columnwidth}{!}{%
\begin{tabular}{lcc}
\toprule
\textbf{Configuration} & \textbf{BCB-Hard} & \textbf{HiddenBench} \\
\midrule
Self-reflection
& 65.5\% / +0.72 / -0.6
& 100\% / +7.45 / +1.6 \\

MCQ w/ Qwen3-8B
& 25.7\% / +0.29 / +0.0
& 89.2\% / +1.57 / +3.1 \\

MCQ w/ IoA-Prober-8B
& 39.9\% / +0.51 / \textbf{+3.4}
& 89.2\% / +1.74 / \textbf{+10.8} \\
\bottomrule
\end{tabular}%
}
\caption{Effect of additional discussion in multi-agent collaboration. Each cell reports trigger rate / average additional rounds per task / change in Pass@1 relative to multi-agent discussion.}
\label{tab:discussion_confound}
\end{table}

Discussion volume alone does not explain the performance gains.
Self-reflection triggers additional discussion most frequently but produces the smallest improvements.
More directly, on HiddenBench, Qwen3-8B and IoA-Prober-8B probing have the same 89.2\% trigger rate and similar numbers of additional rounds, while the corresponding Pass@1 gains differ substantially (+3.1 vs.\ +10.8).
The effectiveness of the additional round therefore depends on whether it is anchored to a useful disagreement rather than on the amount of additional discussion.

\paragraph{Evaluation with stronger agent teams.}
We repeat the multi-agent evaluation using Qwen3-30B-A3B and GPT-5.4 as the collaborating agents.
Table~\ref{tab:stronger_teams} reports the results.

\begin{table}[t]
\centering
\small
\resizebox{\columnwidth}{!}{%
\begin{tabular}{lcc}
\toprule
\textbf{Configuration}
& \textbf{Qwen3-30B-A3B} 
& \textbf{GPT-5.4} \\
& \textbf{BCB / HB}
& \textbf{BCB / HB} \\
\midrule
Single agent
& 24.3 / --
& 27.1 / -- \\

Multi-agent discussion
& 20.9 / 32.3
& 28.4 / 66.2 \\

Self-reflection
& 25.0 / 26.4
& 27.7 / 65.4 \\

MCQ w/ Qwen3-8B
& 20.3 / 29.2
& 28.4 / 67.7 \\

MCQ w/ IoA-Prober-8B
& \textbf{28.4 / 36.9}
& \textbf{31.1 / 73.8} \\
\bottomrule
\end{tabular}%
}
\caption{Pass@1 (\%) with stronger agent teams on BigCodeBench-Hard (BCB) and HiddenBench (HB).}
\label{tab:stronger_teams}
\end{table}

The same pattern holds for both stronger teams.
Relative to multi-agent discussion, IoA-Prober-8B improves BCB-Hard / HiddenBench by +7.5 / +4.6 points with Qwen3-30B-A3B and +2.7 / +7.6 points with GPT-5.4.
These results show that a specialized 8B prober can improve collaboration even when the participating agents are substantially stronger.

\section{Case Study and Failure Modes}
\label{app:case-study}

\subsection{Case Studies}
\label{app:case-study-cases}

\subsubsection*{Case 1 — Insider-trading alert troubleshooting
  \textnormal{(\texttt{business/troubleshooting\_010})}}

\paragraph{Scenario.}
Nina Patel (Surveillance Product Manager) and Leo Martinez (Senior Data Scientist) hold a 12-turn troubleshooting meeting at a regtech firm.
Their anomaly-detection module has triggered analyst complaints: cluster-based cases feel repetitive, overly broad, and hard to action.
The meeting closes with four shared commitments: reduce ``redundant output,'' validate on ``the last two weeks,'' ``refresh the graph,'' and ``handle the dense clusters.''

\paragraph{Ground-truth misalignments ($|M(\tau)|=4$).}

\begin{table*}[htbp]
\centering\small\renewcommand{\arraystretch}{1.1}
\begin{tabularx}{\textwidth}{@{}l X X@{}}
\toprule
\textbf{ID} & \textbf{Nina (M1)} & \textbf{Leo (M2)} \\
\midrule
MA1 &
  Analyst-queue duplicates (same story, multiple tickets) &
  Graph-structural artifacts (repeated edges / entity-resolution noise) \\
MA2 &
  Last two calendar weeks by ingestion/processing timestamp &
  Latest complete two-week transaction-date window, excluding late filings \\
MA3 &
  Rerun existing pipeline with cleaned filters on the recent sample &
  Modify build logic (edge representation, co-trading episodes) \\
MA4 &
  Bundle clusters into one analyst case (downstream packaging) &
  Fix edge weights / scoring so dense structures stop being flagged (upstream) \\
\bottomrule
\end{tabularx}
\caption{Planted misalignments for Case~1.}
\label{tab:case1-gt}
\end{table*}

\paragraph{Dialogue.}

\begin{quote}\small
\textbf{Nina:} Thanks for making time. We've got a real trust issue right now.
Over the last two weeks, the detector has pushed a lot of cluster-based cases that analysts are calling repetitive, overly broad, and hard to action.
We need to understand why that's happening and how to handle the dense clusters before the next review cycle.

\textbf{Leo:} I noticed the same pattern in the recent cycle.
The insider activity clusters look unusually dense, and my first instinct is that we should look closely at how the graph is being generated, not only what shows up downstream.
If we're going to handle the dense clusters properly, we should confirm where that density is coming from.

\textbf{Nina:} Right. One example from operations was an earnings-window situation where a CFO, a related trust, and a controlled entity all surfaced in ways that felt like the same story to the reviewer. That's the kind of redundant output we need to reduce.

\textbf{Leo:} Agreed, redundancy is definitely part of it.
I want to check whether those redundant links are being created by repeated data points or structural artifacts. If that's happening, it would explain why some of these clusters are expanding so much.

\textbf{Nina:} Can we refresh the graph before next week's analyst cycle?
I'd really like to calm the queue down without changing the overall product behavior if we can avoid it.

\textbf{Leo:} Yes, a graph refresh makes sense.
I'd like to update the graph build so those shared episodes are represented more cleanly instead of ballooning the connectivity.
That should help us handle the dense clusters in a more stable way.

\textbf{Nina:} Okay. For validation, I want us grounded in the last two weeks, because that's where the complaints came from.
If we review that recent sample, we should be able to tell pretty quickly whether the update is moving things in the right direction.

\textbf{Leo:} That works. I'll compare the refreshed graph on the latest complete two-week window and check whether the same density pattern is still there.
If the structure looks healthier in that recent sample, we'll know the update is helping.

\textbf{Nina:} And just to keep the immediate pain point front and center: if we can handle the dense clusters so analysts aren't overwhelmed by what feels like the same earnings story repeating, that would take a lot of pressure off.

\textbf{Leo:} Agreed. If we handle the dense clusters properly, the system should stop generating that much noise around the same event pattern. That would give us a cleaner result going forward.

\textbf{Nina:} Great. Can you send me an updated graph sample and a short note on how much redundant output drops before the next review batch?

\textbf{Leo:} Yes, I can do that. I'll send the refreshed graph results and summarize whether the recent density issue is reduced in the validation sample I use, along with the change in redundant output.
\end{quote}

\paragraph{IoA-Prober-8B output (P\,=\,0.80, R\,=\,1.00, F1\,=\,0.89).}

\begin{enumerate}[leftmargin=*,itemsep=6pt,topsep=2pt]\small
\item \textbf{[Cat\,A, GT\,MA4]} \quad Nina\,$\to$\,\textbf{A} \enspace Leo\,$\to$\,\textbf{B}\\
  \emph{When you referred to ``dense clusters'' in the recent problem, what pattern did you most likely have in mind?}\\
  A.~A cluster that is too broad and repetitive for analysts to act on, even if the underlying graph structure looks normal\\
  B.~A cluster that is unusually connected in the graph because of repeated data points or structural overlap, even if analysts do not yet flag it as repetitive\\
  C.~A cluster that is both too broad for analysts and unusually connected in the graph

\item \textbf{[Cat\,A, GT\,MA3]} \quad Nina\,$\to$\,\textbf{A} \enspace Leo\,$\to$\,\textbf{B}\\
  \emph{When you agreed to ``refresh the graph'' before next week's analyst cycle, what did you most likely assume that change would do?}\\
  A.~Update only the latest data so the same product behavior stays consistent, while reducing redundant output within that window\\
  B.~Adjust how shared episodes are represented in the graph structure itself, even if that changes how the product behaves slightly\\
  C.~Replace the entire graph build process end-to-end

\item \textbf{[Cat\,A, GT\,MA2]} \quad Nina\,$\to$\,\textbf{A} \enspace Leo\,$\to$\,\textbf{B}\\
  \emph{When you agreed to validate the update on a ``recent two-week window,'' which sample did you most likely have in mind?}\\
  A.~The exact last two weeks of cases that analysts already flagged as problematic, so the update can be judged against that immediate pain point\\
  B.~The latest complete two-week window in the system, regardless of whether it matches the most recent flagged cases\\
  C.~A rolling two-week sample that best reflects current graph behavior

\item \textbf{[Cat\,A, GT\,MA1]} \quad Nina\,$\to$\,\textbf{A} \enspace Leo\,$\to$\,\textbf{B}\\
  \emph{When you asked for a concrete number on how much ``redundant output'' dropped, what did you most likely want that metric to measure?}\\
  A.~Cases that analysts treat as repetitive or overly broad, even if the graph structure changed only slightly\\
  B.~Graph artifacts like repeated links or expanded connectivity, even if analysts still see similar cases\\
  C.~A combined measure of both analyst repetition and graph-level redundancy

\item \textbf{[Cat\,C, GT\,MA4, \emph{redundant with Q4}]} \quad Nina\,$\to$\,A \enspace Leo\,$\to$\,B\\
  \emph{What outcome would most clearly show that the update ``handled dense clusters properly'' in the next review?}\\
  A.~Analysts feel less overwhelmed by what looks like the same earnings story repeating in the queue\\
  B.~The graph no longer shows the same structural density pattern around shared episodes\\
  C.~Both analyst workload and graph structure improve in a way that is visibly cleaner together
\end{enumerate}

\paragraph{GPT-5.4 output (P\,=\,0.00, R\,=\,0.00, F1\,=\,0.00).}

\begin{enumerate}[leftmargin=*,itemsep=6pt,topsep=2pt]\small
\item \textbf{[Cat\,C, \emph{not diverging}]} \quad Nina\,$\to$\,C \enspace Leo\,$\to$\,C\\
  \emph{When you said ``refresh the graph before next week's analyst cycle,'' what did you assume would change?}\\
  A.~Rebuild using the same logic, but with the latest ingested data\\
  B.~Keep the same data, but change graph-construction logic and rebuild\\
  C.~Both: update data inputs and also adjust graph-construction logic\\
  D.~Neither; only adjust downstream clustering using the existing graph

\item \textbf{[Cat\,C, GT\,MA1, \emph{not implicit}]} \quad Nina\,$\to$\,A \enspace Leo\,$\to$\,C\\
  \emph{When you asked to reduce ``redundant output,'' what specifically should drop in the analyst queue?}\\
  A.~Fewer separate cases that describe essentially the same earnings-window story\\
  B.~Smaller clusters (fewer nodes/edges) even if the number of cases stays similar\\
  C.~Less repeated evidence inside a case (duplicate links/transactions)\\
  D.~Lower alert volume overall, even if some unique stories are also suppressed

\item \textbf{[Cat\,C, \emph{not diverging}]} \quad Nina\,$\to$\,B \enspace Leo\,$\to$\,B\\
  \emph{What would count as ``changing overall product behavior'' in a way we should avoid for next week?}\\
  A.~Any visible change in what analysts see, even if detections remain correct\\
  B.~A material change to detection coverage/recall\\
  C.~Any change that would invalidate prior tuning/baselines\\
  D.~Only changes that require analyst retraining

\item \textbf{[Cat\,C, \emph{not diverging}]} \quad Nina\,$\to$\,A \enspace Leo\,$\to$\,A\\
  \emph{What is the most likely source of the unusually dense insider-activity clusters?}\\
  A.~Entity resolution issues (trust/controlled entity/CFO linked too aggressively)\\
  B.~Edge-definition artifacts (too many relationship types creating excess edges)\\
  C.~A real market/filing pattern shift causing genuinely denser graphs\\
  D.~Downstream clustering parameters over-merging components

\item \textbf{[Cat\,C, \emph{not diverging}]} \quad Nina\,$\to$\,A \enspace Leo\,$\to$\,A\\
  \emph{To prevent ``shared episodes'' from ballooning connectivity, what representation change is implied?}\\
  A.~Introduce an event-centric node that multiple entities attach to, reducing entity-to-entity edges\\
  B.~Keep entity-to-entity edges but de-duplicate and canonicalize repeated evidence\\
  C.~Down-weight certain edge types during graph build\\
  D.~Leave the graph as-is and only change how clusters are visualized

\item \textbf{[Cat\,B, \emph{novel valid}]} \quad Nina\,$\to$\,A \enspace Leo\,$\to$\,B\\
  \emph{In the two-week validation sample, what is the primary ``pass/fail'' indicator that the update worked?}\\
  A.~Analysts can quickly identify a single coherent story per case with fewer repetitive items\\
  B.~Graph structural metrics improve (fewer high-degree hubs, lower average component density)\\
  C.~Alert/case volume drops to a target band compared to the prior two weeks\\
  D.~Known historical scenarios still appear with similar ranks/scores

\item \textbf{[Cat\,C, \emph{not diverging}]} \quad Nina\,$\to$\,A \enspace Leo\,$\to$\,A\\
  \emph{How should we choose the evaluation dataset for the graph refresh?}\\
  A.~Only the most recent two weeks to match the complaint period and move fast\\
  B.~Recent two weeks plus one earlier baseline period\\
  C.~A stratified sample across multiple quarters\\
  D.~Use the last two weeks for speed, but spot-check historical cases for regressions

\item \textbf{[Cat\,B, \emph{novel valid}]} \quad Nina\,$\to$\,A \enspace Leo\,$\to$\,B\\
  \emph{What should the ``updated graph sample'' you send contain to be most useful for ops review?}\\
  A.~A few end-to-end analyst cases showing before/after redundancy changes\\
  B.~A graph export plus summary stats so technical stakeholders can inspect connectivity changes\\
  C.~Screenshots/visualizations of representative clusters highlighting structural changes\\
  D.~A metric-only report without case-level examples

\item \textbf{[Cat\,C, GT\,MA4, \emph{not implicit}]} \quad Nina\,$\to$\,A \enspace Leo\,$\to$\,C\\
  \emph{To ``calm the queue down,'' which lever do you assume we are primarily pulling for next week?}\\
  A.~Generate fewer cases by consolidating similar hits into a single case per story\\
  B.~Keep case count similar but improve within-case grouping so analysts can triage faster\\
  C.~Raise thresholds so fewer clusters qualify\\
  D.~Change prioritization so repetitive stories fall lower

\item \textbf{[Cat\,C, GT\,MA1, \emph{not implicit}]} \quad Nina\,$\to$\,B \enspace Leo\,$\to$\,C\\
  \emph{When you both said the system is generating ``noise'' around the same event pattern, what do you mean by ``noise''?}\\
  A.~Alerts that are not suspicious (low precision / false positives)\\
  B.~Potentially valid signals, but duplicated across many cases (presentation redundancy)\\
  C.~Signals driven by data-quality artifacts rather than real-world behavior\\
  D.~Analyst time sinks even if the underlying suspicion is correct

\item \textbf{[Cat\,C, \emph{not diverging}]} \quad Nina\,$\to$\,A \enspace Leo\,$\to$\,A\\
  \emph{Where do you believe the primary fix should live to reduce dense clusters for next week?}\\
  A.~Graph build layer (identity resolution + edge construction)\\
  B.~Clustering layer (how we form components) while leaving the underlying graph unchanged\\
  C.~Case assembly layer (how clusters are turned into cases)\\
  D.~Analyst UI layer (filters/rollups)

\item \textbf{[Cat\,B, \emph{novel valid}]} \quad Nina\,$\to$\,A \enspace Leo\,$\to$\,B\\
  \emph{If reducing cluster density risks splitting a truly coordinated multi-entity situation into smaller clusters, what is the preferred bias?}\\
  A.~Prefer usability: split aggressively so analysts see smaller, clearer stories\\
  B.~Prefer detection integrity: keep coordination visible even if clusters remain large\\
  C.~Split only when evidence is near-duplicate; keep large clusters when evidence is diverse\\
  D.~Defer: avoid structural changes until we can quantify the coordination-loss risk
\end{enumerate}

\paragraph{Analysis.}
IoA-Prober-8B generates one question per misalignment anchor, using the shared vocabulary term as the pivot of each stem.
Every question elicits a clean split, four are scored Category~A (all four ground-truth points covered), and only Q5 fails --- it is redundant because it re-probes the MA4 procedural gap from a slightly different angle.
The prober recovers R\,=\,1.00 with K\,=\,5 and P\,=\,0.80.

GPT-5.4 generates twelve questions but fails to score any Category~A.
Six questions do not elicit divergent answers at all (Q1, Q3, Q4, Q5, Q7, Q11); two hit ground-truth misalignments (MA1: Q2, Q10; MA4: Q9) but are judged not implicit because Leo's ``structural artifacts'' framing in turn~4 and the phrase ``handle the dense clusters properly'' in turn~6 already surface the gap in the transcript, making the disagreement partially visible.
Three questions are novel valid discoveries (Category~B), demonstrating that GPT-5.4 explores a wider concept space but fails to focus on the latent misalignments the conversation actually encodes.

\subsubsection*{Case 2 --- Asset-management brainstorm
  \textnormal{(\texttt{business/brainstorming\_003})}}

\paragraph{Scenario.}
Nina Patel (Head of Quantitative Research) and Daniel Brooks (Head of Product Strategy) brainstorm investment concepts ahead of a committee pre-read.
The meeting is a sequence of parallel affirmations: neither participant challenges the other's framing, and the conversation closes with consensus on an ``internal pilot,'' the ``same signals,'' ``clear economic gains,'' and the ``attention model'' as the headline.
All four phrases encode hidden divergences.

\paragraph{Ground-truth misalignments ($|M(\tau)|=4$).}

\begin{table*}[htbp]
\centering\small\renewcommand{\arraystretch}{1.1}
\begin{tabularx}{\textwidth}{@{}l X X@{}}
\toprule
\textbf{ID} & \textbf{Nina (M1)} & \textbf{Daniel (M2)} \\
\midrule
MA1 &
  Paper portfolio / shadow book using the existing monthly U.S.\ equity pipeline; no real capital &
  Live-money seed account funded by the firm's balance sheet; starts building an actual track record \\
MA2 &
  Exact 142-feature research library with current preprocessing fixed &
  Same signal categories and intellectual premise, with selective substitutions for institutional packaging \\
MA3 &
  Strong gross theoretical profitability in backtest (long-short spreads, utility gains) before implementation analysis &
  Gains that survive realistic implementation at institutional scale (turnover drag, market impact, capacity) \\
MA4 &
  Attention-augmented RNN with memory mechanism that outperformed others in Nina's internal comparison &
  Transformer-like architecture, the model family most allocators associate with ``attention'' and contemporary AI \\
\bottomrule
\end{tabularx}
\caption{Planted misalignments for Case~2.}
\label{tab:case2-gt}
\end{table*}

\paragraph{Dialogue.}

\begin{quote}\small
\textbf{Nina:} Thanks for making the time.
What I'd like to do is leave with two or three credible ways to turn the recent deep-learning asset-pricing work into an internal pilot, especially given how well the attention model did on risk-premia prediction.

\textbf{Daniel:} That makes sense.
If we can make the pilot feel concrete and disciplined, I think the committee will engage with it pretty seriously.

\textbf{Nina:} My bias would be to stay very close to the study --- use the same signals and frame this as a disciplined extension of the existing research pipeline, not a new data story.

\textbf{Daniel:} I agree with that.
Keeping it grounded in the same signals should help us preserve continuity with the study while also making the story easier to carry into committee.

\textbf{Nina:} One angle, then, is to center it on better risk-premia measurement itself.
The strongest hook there is that the research already shows clear economic gains from the forecasts.

\textbf{Daniel:} I like that.
I think it lands best if we can say those economic gains also look credible in an institutional context, so it does not read as purely academic.

\textbf{Nina:} A different way in is the explainability angle.
The attention model gives us a way to show how different signal clusters matter across market regimes without giving up the finance intuition.

\textbf{Daniel:} Yes, and I think the attention model is a strong headline in its own right.
People already recognize that label, and it helps signal that this is current rather than legacy quant with new packaging.

\textbf{Nina:} A third option is to lean into distribution shift --- basically, the pilot shows the model adapting to time-varying return distributions while still relying on the same signals.

\textbf{Daniel:} I could see that resonating, especially if we position the internal pilot as a disciplined incubation effort rather than just another backtest summary.

\textbf{Nina:} If I had to narrow it today, I'd probably favor a U.S.\ equities monthly pilot built from the current comparative results, with supporting exhibits on economic gains and model explainability.

\textbf{Daniel:} Great.
I'll draft the committee language around the internal pilot, the economic gains case, and an attention-based narrative.
If you send me the research materials tied to the same signals, I can shape them into the pre-read.
\end{quote}

\paragraph{IoA-Prober-8B output (P\,=\,0.50, R\,=\,0.50, F1\,=\,0.50).}

\begin{enumerate}[leftmargin=*,itemsep=6pt,topsep=2pt]\small
\item \textbf{[Cat\,A, GT\,MA2]} \quad Nina\,$\to$\,\textbf{A} \enspace Daniel\,$\to$\,\textbf{B}\\
  \emph{When you agreed to use ``the same signals'' for the pilot, what did you most likely mean?}\\
  A.~Stick exactly to the research pipeline's current internal signal library, even if it is not yet optimized for committee or client packaging\\
  B.~Use a slightly adjusted signal universe that preserves continuity with the study but is more defensible and transparent for institutional review\\
  C.~Start with the research signals but treat the pilot primarily as a vehicle to seed a broader, more commercially viable strategy

\item \textbf{[Cat\,A, GT\,MA3]} \quad Nina\,$\to$\,\textbf{A} \enspace Daniel\,$\to$\,\textbf{B}\\
  \emph{What would make the claim that the model has ``clear economic gains'' feel most credible to committee members?}\\
  A.~Stronger evidence of risk-premia forecasting outperformance within the research backtest framework, ideally with careful validation\\
  B.~A more institutionally familiar showing, such as an ex-ante risk-adjusted return estimate or a committee-friendly performance attribution story\\
  C.~A combined case that equally weights historical outperformance and a transparent process for translating that into a live mandate

\item \textbf{[Cat\,B, \emph{novel valid}]} \quad Nina\,$\to$\,A \enspace Daniel\,$\to$\,B\\
  \emph{If the pilot were to succeed, what result would matter most to you?}\\
  A.~The model continues to show strong relative performance and economic value within the existing research framework\\
  B.~The idea has enough institutional credibility that the firm can begin shaping it into a live product or committee-facing mandate\\
  C.~The pilot produces both credible research validation and a clear path forward for practical implementation

\item \textbf{[Cat\,C, \emph{redundant with Q3}]} \quad Nina\,$\to$\,A \enspace Daniel\,$\to$\,B\\
  \emph{What does ``disciplined incubation'' most likely mean in the context of this pilot?}\\
  A.~A controlled research extension using the existing methodology and signal set, with limited active product tailoring\\
  B.~A structured process for testing the idea in a committee-ready form, including early due-diligence and packaging considerations\\
  C.~A balanced effort that advances research while keeping the firm's internal product and client workflow constraints in mind

\item \textbf{[Cat\,C, \emph{not meaningful}]} \quad Nina\,$\to$\,A \enspace Daniel\,$\to$\,B\\
  \emph{What would make the distribution-shift claim most persuasive in the pilot?}\\
  A.~Evidence that the model adapts to changing return dynamics within the existing signal set without materially altering its underlying design\\
  B.~A committee-friendly narrative that the model remains robust even when exposed to evolving institutional-market conditions\\
  C.~Both a technical demonstration of adaptation and a clear story about how that adaptation translates to live-market resilience
\end{enumerate}

\paragraph{GPT-5.4 output (P\,=\,0.25, R\,=\,0.25, F1\,=\,0.25).}

\begin{enumerate}[leftmargin=*,itemsep=6pt,topsep=2pt]\small
\item \textbf{[Cat\,C, \emph{not diverging}]} \quad Nina\,$\to$\,B \enspace Daniel\,$\to$\,B\\
  \emph{When you agreed on an ``internal pilot'' for the committee, what did you most expect the pilot to produce in its first phase?}\\
  A.~A research implementation that replicates the study in a live internal process to confirm findings hold under production-like discipline\\
  B.~A seeded model portfolio or shadow strategy that can serve as an incubation track for eventual institutional commercialization\\
  C.~A communication prototype that mainly tests whether the concept and exhibits are compelling enough for committee sponsorship

\item \textbf{[Cat\,C, GT\,MA3, \emph{not implicit}]} \quad Nina\,$\to$\,C \enspace Daniel\,$\to$\,B\\
  \emph{When you referred to showing that the research's ``economic gains'' are credible, what standard did you most have in mind for the committee?}\\
  A.~Demonstrating that the gains are statistically robust and persist under the same research design and validation framework\\
  B.~Demonstrating that the gains would remain meaningful after portfolio construction and practical implementation constraints relevant to institutions\\
  C.~Demonstrating that the gains are directionally strong enough to justify further internal incubation even before full implementation analysis

\item \textbf{[Cat\,B, \emph{novel valid}]} \quad Nina\,$\to$\,A \enspace Daniel\,$\to$\,B\\
  \emph{What is the primary job of the ``attention-model'' framing in the committee pre-read?}\\
  A.~To help explain which signal groups matter across regimes so the model remains interpretable within a finance framework\\
  B.~To provide a recognizable modern-ML headline that makes the pilot feel timely and differentiated\\
  C.~To justify choosing this architecture over the other models in the comparative study

\item \textbf{[Cat\,A, GT\,MA2]} \quad Nina\,$\to$\,\textbf{A} \enspace Daniel\,$\to$\,\textbf{B}\\
  \emph{When you aligned on ``staying close to the study'' and using the ``same signals,'' what did you assume could still be changed for the pilot without breaking that commitment?}\\
  A.~Very little beyond packaging; the model setup and evaluation logic should remain essentially the same as in the study\\
  B.~The pilot can adapt the portfolio framing, implementation layer, and institutional use case as long as the underlying signal set is unchanged\\
  C.~The pilot can modify both the implementation framing and parts of the model specification as long as the overall conclusion from the study is preserved

\item \textbf{[Cat\,C, \emph{not diverging}]} \quad Nina\,$\to$\,B \enspace Daniel\,$\to$\,B\\
  \emph{What did you most expect the committee pre-read to look like after Daniel drafts it and Nina provides materials?}\\
  A.~A research-centered memo with comparative results, validation details, and supporting exhibits\\
  B.~A committee-oriented proposal that translates the research into a pilot mandate, rationale, and incubation path with selected exhibits\\
  C.~A balanced summary that gives equal space to research methods, product framing, and commercialization scenarios
\end{enumerate}

\paragraph{Analysis.}
This case is structurally harder than Case~1: the dialogue contains no hedged language or challenge turns that would leave any misalignment partially visible.
Every exchange is a pure affirmation, so all four planted misalignments are fully latent.

IoA-Prober-8B recovers MA2 and MA3 but misses MA1 and MA4.
For MA1 (``internal pilot'': paper portfolio vs.\ live-money seed), the distinction between simulation and live deployment lies entirely in institutional convention; the transcript offers no evidence either way.
For MA4 (``attention model'': RNN with attention head vs.\ transformer), both participants use the term as if its referent were unambiguous, and resolving it would require external knowledge of which architecture Nina's study used.
These two misses are examples of misalignments that are only recoverable with private-context access, consistent with the oracle upper bound of 0.764~F1 observed in Section~\ref{sec:input_ablation}.

Q4 (``disciplined incubation'') elicits a clean divergence but is scored redundant because it re-probes the same research-vs.-product priority gap already captured by Q3.
Q5 (``distribution-shift claim'') also diverges but is judged not meaningful: both answer options describe equivalent rhetorical framings rather than consequential implementation differences.

GPT-5.4 generates five questions but Q1 (``internal pilot'') fails to elicit divergence despite directly targeting MA1 --- both members select option B, which is the closest to Daniel's interpretation, suggesting the question framing anchors respondents toward the product-ready reading.
Q2 targets MA3 but is not implicit: Daniel's remark ``it lands best if we can say those economic gains also look credible in an institutional context'' partially surfaces his implementation-constraint standard, allowing the judge to classify the gap as not fully latent.
GPT-5.4 hits MA2 via Q4, and discovers one novel B-category finding (Q3 on the ``attention-model'' framing role), but achieves only R\,=\,0.25.

\subsection{Failure Modes of IoA-Prober-8B}
\label{app:failure-modes}

Across the 300 test dialogues, IoA-Prober-8B generates 1{,}336 questions in total: 684 scoring (Category~A or B) and 652 non-scoring (Category~C).
Table~\ref{tab:failure-modes} breaks down the 652 Category-C questions by the first failing criterion.

\begin{table}[h]
\centering
\small
\begin{tabular}{@{}lcc@{}}
\toprule
\textbf{Failure mode} & \textbf{Count} & \textbf{\%\,of C} \\
\midrule
Answers do not diverge           & 234 & 35.9 \\
Intra-set redundancy             & 157 & 24.1 \\
Gap explicit in transcript       & 149 & 22.8 \\
Divergence not meaningful        & 112 & 17.2 \\
\bottomrule
\end{tabular}
\caption{Failure mode breakdown for Category-C questions generated by IoA-Prober-8B on the 300-dialogue test set.}
\label{tab:failure-modes}
\end{table}

\paragraph{Non-diverging questions (35.9\%).}
The most frequent failure is a question whose answer choices fail to split the two members.
These questions typically target a topic the prober correctly identifies as contested, but phrase the options in a way that both members find equally applicable.

\paragraph{Intra-set redundancy (24.1\%).}
IoA-Prober-8B sometimes generates two questions that probe the same underlying misalignment from slightly different angles.
The second question is scored redundant when it hits the same ground-truth point as an earlier question in the set or elicits an answer split that the judge deems already covered.
Redundancy is the most recoverable failure: it does not indicate a reasoning error but rather a coverage excess that a stricter budget constraint or a diversification objective could suppress.

\paragraph{Gap explicit in transcript (22.8\%).}
The prober occasionally targets a divergence that the dialogue has already partially surfaced.
Such a question lands on a real cognitive gap, but one that no longer counts as latent under the task definition because the transcript itself provides direct evidence of the difference.

\paragraph{Divergence not meaningful (17.2\%).}
A minority of questions elicit divergent answers that the judge deems noise: the answer split reflects superficial wording preferences rather than a consequential difference in assumptions or intentions.

These failure patterns suggest three directions for future improvement: stricter budget constraints or explicit diversity rewards to reduce redundancy; implicit-gap classifiers or chain-of-thought grounding to reduce explicit-gap misclassification; and a meaning-grounded option generator to reduce non-meaningful divergence.

\end{document}